\documentclass[10pt,letterpaper]{article}
\usepackage[T1]{fontenc}
\usepackage{configuration/iclr2027_conference,times}
\iclrfinalcopy
\usepackage{configuration/arxiv}
\usepackage{xcolor}
\definecolor{darkred}{RGB}{150,0,0}
\usepackage[hypertexnames=false]{hyperref}
\hypersetup{
  colorlinks=true,
  linkcolor=darkred,
  citecolor=darkred,
  urlcolor=darkred
}
\usepackage{url}
\usepackage{amsmath,amssymb,amsthm,booktabs,array,tabularx}
\usepackage{graphicx}
\usepackage{float,placeins}
\usepackage{pgfplots,pgfplotstable}
\pgfplotsset{compat=1.18}
\usepackage{algorithm,algpseudocode}
\usepackage{capt-of,listings}
\usepackage{colortbl}
\usepackage{enumitem}
\usepackage{xspace}

\newcommand{\darkredtext}[1]{{\color{darkred}#1}}
\newcommand{\secref}[1]{\hyperref[#1]{\darkredtext{Sec.~\ref*{#1}}}}
\newcommand{\thmref}[1]{\hyperref[#1]{\darkredtext{Thm.~\ref*{#1}}}}
\newcommand{\propref}[1]{\hyperref[#1]{\darkredtext{Prop.~\ref*{#1}}}}
\newcommand{\assumpref}[1]{\hyperref[#1]{\darkredtext{Assump.~\ref*{#1}}}}
\newcommand{\remarkref}[1]{\hyperref[#1]{\darkredtext{Rem.~\ref*{#1}}}}
\newcommand{\algxref}[1]{\hyperref[#1]{\darkredtext{Alg.~\ref*{#1}}}}
\newcommand{\figref}[1]{\hyperref[#1]{\darkredtext{Fig.~\ref*{#1}}}}
\newcommand{\tabref}[1]{\hyperref[#1]{\darkredtext{Tab.~\ref*{#1}}}}
\newcommand{\appref}[1]{\hyperref[#1]{\darkredtext{App.~\ref*{#1}}}}

\newcommand{\sg}{\operatorname{sg}}
\newcommand{\NextGAN}{\textsc{NextGAN}\xspace}
\newcommand{\Dnext}{\cD_{\mathrm{Next}}}
\newcommand{\E}{\mathbb E}
\newcommand{\R}{\mathbb{R}}
\newcommand{\dm}{\mathrm{d}}
\newcommand{\cD}{\mathcal{D}}
\newcommand{\cF}{\mathcal{F}}
\newcommand{\cG}{\mathcal{G}}
\newcommand{\cH}{\mathcal{H}}
\newcommand{\cL}{\mathcal{L}}
\newcommand{\cP}{\mathcal{P}}
\newcommand{\cV}{\mathcal{V}}
\renewcommand{\aa}{\mathbf{a}}
\newcommand{\bb}{\mathbf{b}}
\newcommand{\zero}{\mathbf{0}}
\newcommand{\xx}{\mathbf{x}}
\newcommand{\yy}{\mathbf{y}}
\newcommand{\zz}{\mathbf{z}}
\newcommand{\cc}{\mathbf{c}}
\newcommand{\vv}{\mathbf{v}}
\newcommand{\ww}{\mathbf{w}}
\newcommand{\mX}{\mathbf{X}}
\newcommand{\mC}{\mathbf{C}}
\newcommand{\mY}{\mathbf{Y}}
\newcommand{\mtheta}{\boldsymbol{\theta}}
\newcommand{\mphi}{\boldsymbol{\phi}}
\newcommand{\mDelta}{\boldsymbol{\Delta}}
\newcommand{\mg}{\boldsymbol{g}}
\newcommand{\mf}{\boldsymbol{f}}
\newcolumntype{Y}{>{\raggedright\arraybackslash}X}
\newtheorem{theorem}{Theorem}
\newtheorem{assumption}{Assumption}
\newtheorem{proposition}{Proposition}
\theoremstyle{remark}
\newtheorem{remark}{Remark}
\definecolor{nextganaccent}{RGB}{52,60,58}
\definecolor{nextganpale}{RGB}{245,243,238}
\definecolor{videobaselinelabel}{RGB}{145,145,145}
\definecolor{videonextganlabel}{RGB}{200,35,45}
\newcommand{\best}[1]{\textbf{#1}}

\newif\iffinal
\finaltrue

\title{NextGAN: Towards Simple and Scalable GAN Training for Image and Video Generation}
\author{\begin{minipage}{\textwidth}\raggedright\normalsize\bfseries
Xinhao Lou\textsuperscript{1,$\ast$}, Dongye Xie\textsuperscript{2,$\ast$}, Peng Sun\textsuperscript{1,4,$\ast$}, Chenyi Zhuang\textsuperscript{3}, Kaicheng Yu\textsuperscript{1,$\dagger$}, Tao Lin\textsuperscript{1,$\dagger$}\\[4pt]
\normalfont\small\textsuperscript{1}Westlake University\quad\textsuperscript{2}University of Cambridge\quad\textsuperscript{3}Independent Researcher\quad\textsuperscript{4}Zhejiang University\\[2pt]
\footnotesize $\ast$ Equal contribution.\quad $\dagger$ Corresponding authors.
\end{minipage}}
\date{}
\begin{document}
\maketitle
\begingroup
\linespread{1.09}\selectfont

\begin{abstract}
Adversarial training enables one- and few-step image and video generation, but conventional adversarial losses supervise critic values without directly specifying the gradients used to update the generator.
We present \NextGAN, which trains a scalar critic by regressing its input gradient onto displacements between independently paired reference and generated features at interpolated locations in frozen representation spaces.
For a fixed generator, this regression is equivalent to an adversarial objective with a gradient-energy penalty, while the generator retains the standard potential-based GAN update.
The same objective adapts pretrained image and video generators with up to 14B parameters, without auxiliary losses, online teacher scores, or trajectories.
On ImageNet $256\times256$, \NextGAN achieves one-step FIDs of 2.87 with pixel-space JiT-B and 1.54 with latent-space DiT-XL/1; its adversarial form further reaches 1.36.
For text-to-video generation, a single trained Wan2.1-1.3B model supports one-, two-, and four-step sampling, achieving the highest VBench total scores among the compared 1.3B methods at one and two steps.
At two steps, \NextGAN scores 83.13, exceeding rCM and AnyFlow by 1.25 and 0.47 points, respectively.
Code is available at \url{https://github.com/LINs-lab/NextGAN.git}.
\end{abstract}

\begin{figure}[!b]
\centering
\includegraphics[width=\linewidth]{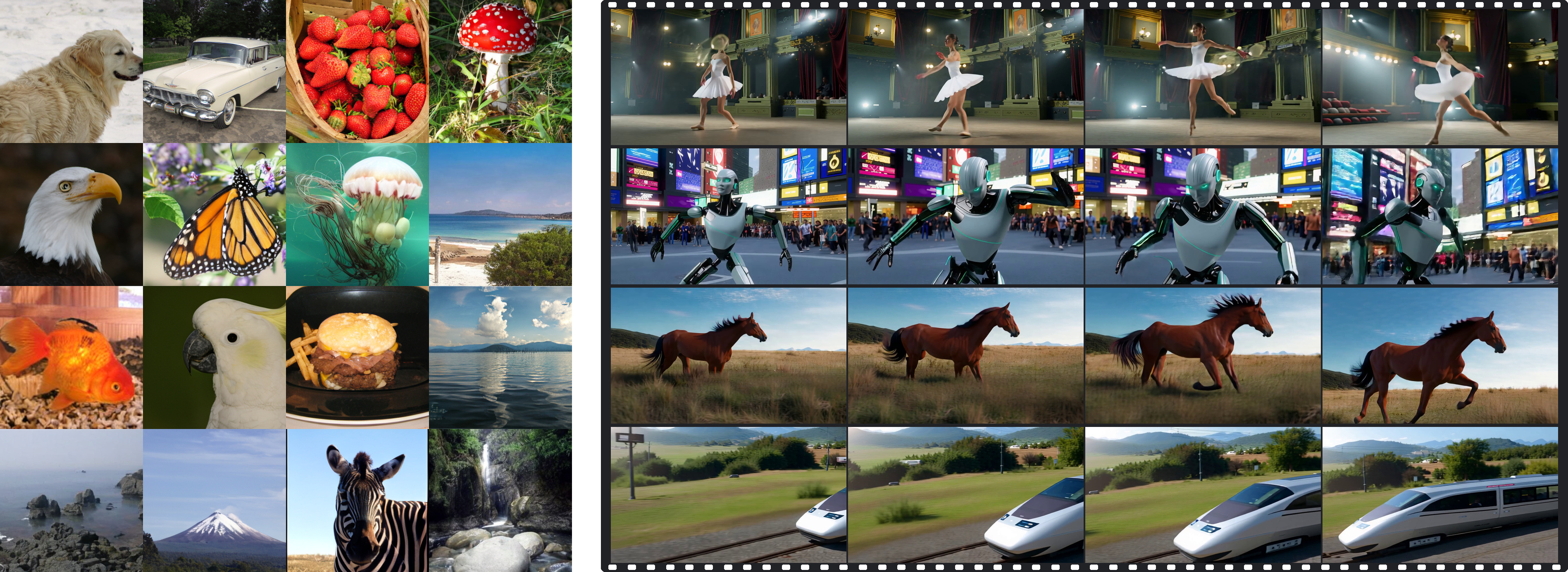}
\small\caption{\textbf{Qualitative samples from \NextGAN.} \emph{Left:} 1-NFE class-conditional samples on ImageNet $256\times256$. \emph{Right:} four frames from each of four generated videos.}
\label{fig:qualitative-overview}
\end{figure}

\begin{figure}[!t]
\centering
\includegraphics[width=\linewidth]{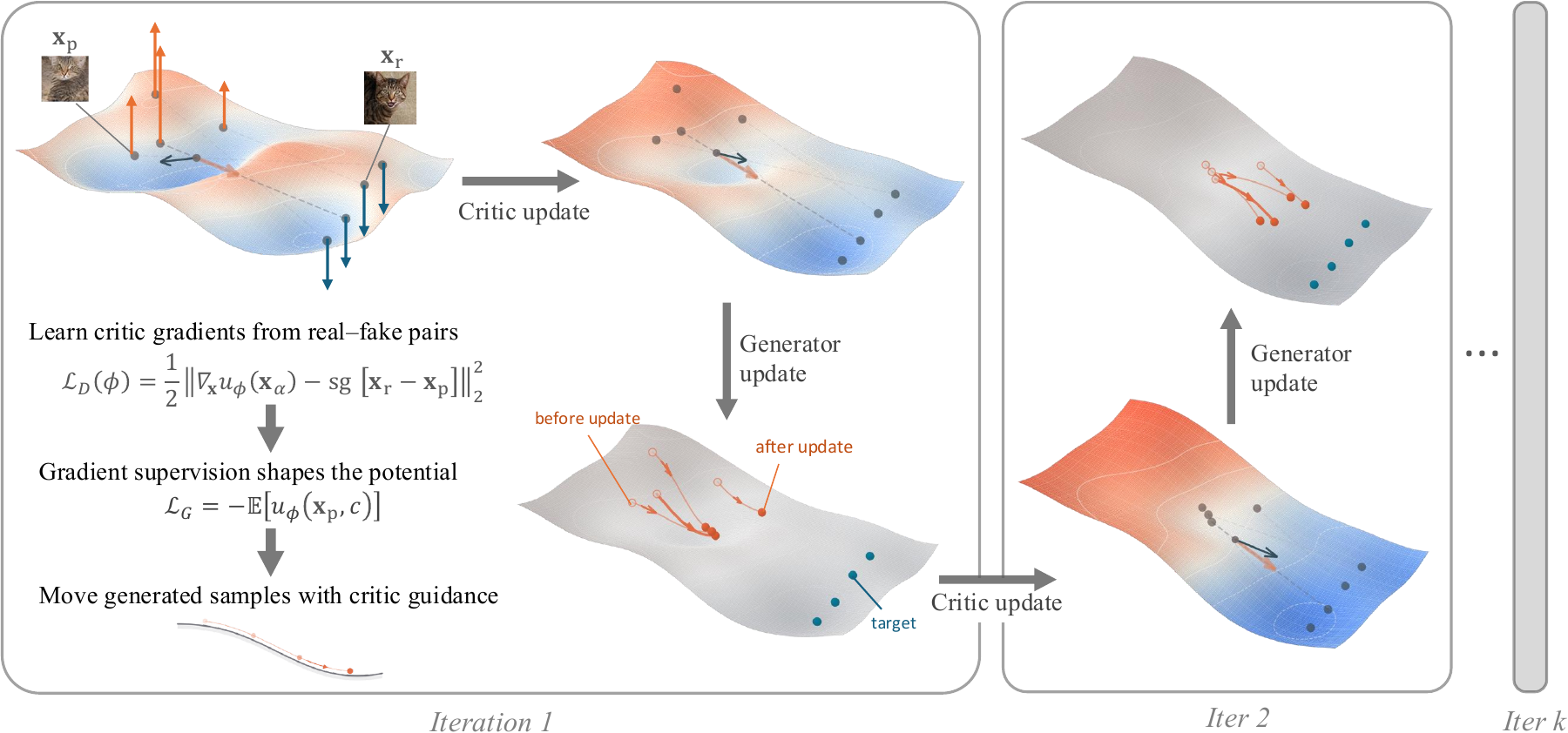}
\small\caption{\textbf{How \NextGAN learns to guide the generator.} $\xx_{\mathrm{p}}$ and $\xx_{\mathrm{r}}$ are generated and real features; $\xx_\alpha$ lies between them. First, the critic loss $\cL_D$ trains $\nabla_{\xx}u_{\mphi}$ to predict their displacement, with endpoints fixed ($\sg$: stop gradient). Then, the generator loss $\cL_G$ moves generated features uphill in the fixed potential $u_{\mphi}$. The surface depicts $-u_{\mphi}$, so this motion appears downhill. These two updates alternate throughout training.}
\label{fig:nextgan}
\end{figure}

\section{Introduction}
\label{sec:introduction}
GANs generate samples in a single forward pass~\citep{goodfellow2014generative}, whereas diffusion and flow models repeatedly evaluate large networks during sampling~\citep{song2021score,lipman2023flow}. Adapting pretrained generators through adversarial training can therefore enable one- or few-step generation from reference samples, making the critic's learning signal central.

The generator consumes the critic's input gradient, whereas conventional adversarial objectives supervise differences in critic values on real and generated samples. The resulting update direction depends on the objective, parameterization, and regularization; successful separation alone need not provide useful gradients~\citep{arjovsky2017wasserstein}. Penalties on gradient magnitude constrain this signal~\citep{gulrajani2017improved}, but do not supply a direction target. We investigate such a target using displacements between reference samples and the current generator's outputs.

We present \NextGAN, which trains a scalar critic by regressing its input gradient onto real--generated displacements in a frozen feature space. As illustrated in \figref{fig:nextgan}, independent pairs supervise the gradient along segments joining their features, and the generator then ascends the learned potential at its own features. Frozen representers provide complementary comparison spaces during training, while inference evaluates only the generator. \figref{fig:qualitative-overview} previews the resulting one-step image samples and generated video sequences.

Independent pairing makes each displacement noisy, but regression aggregates targets by interpolation location and condition. Over an idealized linear potential class, the optimal critic gradient is the conservative projection of this conditional mean displacement. When the conditional feature distributions match, exchanging the endpoints reverses the displacement without changing the interpolation law, so the mean field vanishes. Random pairs therefore supervise a distribution-dependent field without assigning samples fixed destinations; under the assumptions in \propref{prop:population-field}, a sufficiently rich potential class makes the induced discrepancy zero if and only if the conditional feature distributions match.

The regression also has an exact adversarial interpretation: for a fixed generator, averaging over uniform interpolation points yields a real--generated potential gap with a gradient-energy penalty. This connects the construction to Sobolev GAN and to TBSM's conservative displacement construction, which yields the same critic objective~\citep{mroueh2018sobolev,sun2026threebody}. We develop this supervision for standard generator updates at generated endpoints and evaluate both regression and adversarial implementations. Their population objectives are equivalent, although their finite-minibatch updates can differ.

We evaluate \NextGAN through controlled mode-coverage studies, image-generation experiments, and adaptation of Wan2.1 video backbones with up to 14B parameters. The controlled study shows balanced coverage of eight modes, and image ablations support combining complementary representers. On ImageNet $256\times256$, regression achieves 1.54 FID with a latent DiT-XL/1 and 2.87 with a pixel-space JiT-B, each in one evaluation without guidance; the adversarial form achieves 1.36 with DiT-XL/1. For video, the same 1.3B weights support one-, two-, and four-step sampling, reaching VBench total scores of 80.80 and 83.13 at one and two steps, respectively, the highest among the compared 1.3B methods at those budgets (\tabref{tab:vbench}). All large-scale experiments adapt pretrained generators without online teacher scores or trajectories; the video reference samples were generated offline by Wan2.1-14B.

Our contributions are:
\begin{enumerate}[label=(\alph*), nosep, leftmargin=16pt]
    \item \textbf{Displacement-supervised critic training.} We train critic gradients from independent real--generated pairs in frozen feature spaces and use the learned potentials for standard generator updates.
    \item \textbf{A distributional interpretation.} We characterize the optimal field as a projected conditional mean displacement, establish its exact population adversarial dual, and give conditions under which the discrepancy detects conditional feature-distribution mismatch.
    \item \textbf{Image and video generation at scale.} Controlled mode-coverage and representer studies accompany one-step pixel- and latent-space image generation and few-step video adaptation at 1.3B and 14B parameters, without online teacher queries.
\end{enumerate}

\section{Related Work}
\label{sec:related}
\paragraph{Adversarial discrepancies.}
GANs train a generator against a critic that separates data from generated samples~\citep{goodfellow2014generative}; later work changes the discrepancy or regularizes the critic with Lipschitz constraints, gradient penalties, and feature projections~\citep{arjovsky2017wasserstein,gulrajani2017improved,miyato2018spectral,sauer2021projected}. Sobolev GAN is closest in form because it penalizes critic-gradient energy~\citep{mroueh2018sobolev}. \NextGAN differs in the supervision: the gradient itself is regressed onto real--fake displacements, and the resulting value is the corresponding negative-Sobolev discrepancy (Eq.~\eqref{eq:distribution-objective}).

\paragraph{Flow matching and direct distribution dynamics.}
Flow matching regresses a velocity along a prescribed noise--data path~\citep{lipman2023flow}. Direct distribution-dynamics methods instead construct updates from the current model and data distributions: Drifting uses batch interactions, while Three-Body Scattering Modeling (TBSM) uses constant-size real--generated--generated interactions and tracked regression targets~\citep{deng2026drifting,sun2026threebody}. TBSM's design map also contains a GAN-like displacement corner at $\rho=1,\lambda=0$: replacing unit bearings by real--generated displacements and constraining its tracker to be conservative yields the same path-regularized critic objective as our field regression.

\noindent We call this configuration the \emph{TBSM GAN corner}; its generator query differs from the endpoint update used by \NextGAN. \NextGAN uses independent real--generated pairs, marginalizes the path time, and constrains the learned field to be conservative so that it integrates to a scalar potential.

\paragraph{Few-step generation from pretrained models.}
Distribution-matching distillation uses teacher scores or trajectories~\citep{yin2024improved,rcm2025,anyflow2026}, while adversarial post-training introduces a discriminator. \NextGAN constructs its target from reference data and current generator samples, without querying teacher scores or trajectories during training. Throughout the paper, \emph{teacher-free} refers specifically to the absence of such online teacher queries.

\section{NextGAN}
\label{sec:method}

We formalize \NextGAN by defining displacement targets from current real--generated pairs, deriving their adversarial dual, and specifying alternating updates (\figref{fig:nextgan}).

\subsection{Comparison space and random-pair supervision}
Let $(\bar{\xx}_{\mathrm{r}},\cc)\sim p_{\mathrm{data}}(\bar{\xx},\cc)$ be a real sample--condition pair, draw $\zz\sim p(\zz)$ independently of this pair, and let $\bar{\xx}_{\mathrm{p}}=\mg_{\mtheta}(\zz,\cc)$ be a generated sample. For a frozen differentiable representation map $\mf$, define the comparison vectors $\xx_{\mathrm{r}}=\mf(\bar{\xx}_{\mathrm{r}},\cc)$ and $\xx_{\mathrm{p}}=\mf(\bar{\xx}_{\mathrm{p}},\cc)$; an image representer may simply ignore $\cc$. We write their conditional laws as $Q(\cdot\mid\cc)$ and $P_{\mtheta}(\cdot\mid\cc)$, respectively. The endpoints are independent conditional on $\cc$. Define their displacement and the segment between them:
\begin{equation}
    \mDelta=\xx_{\mathrm{r}}-\xx_{\mathrm{p}},\qquad
    \xx_\alpha=(1-\alpha)\xx_{\mathrm{p}}+\alpha\xx_{\mathrm{r}},
    \qquad \alpha\sim\mathcal{U}(0,1).
    \label{eq:corridor}
\end{equation}
All expectations use the above sampling rule with independent $\alpha$. We call the law of $(\xx_\alpha,\cc)$ the \emph{corridor distribution} $\mu_{\mtheta}$. Each displacement is a noisy observation of how the generated distribution differs from the data distribution; the pairs are not assumed to be an optimal-transport coupling.

The critic is a scalar potential $u_{\mphi}(\xx,\cc)$, with trainable parameters $\mphi$ and guidance field $\nabla_{\xx}u_{\mphi}$. Using a potential makes the field compatible with a standard GAN generator update and permits the line-integral identity below. The representer $\mf$ remains frozen.

\subsection{Learning a displacement field}
Conventional GAN objectives supervise critic values on real and generated samples. Here each pair supplies a vector target for the critic gradient, specifying both its direction and magnitude. With the generator fixed, the unit-weight regression objective is
\begin{equation}
    \cL_{\mathrm{field}}(\mphi;\mtheta)
    =\frac{1}{2}\,\E\left[
    \left\|\nabla_{\xx}u_{\mphi}(\xx_\alpha,\cc)-\mDelta\right\|_2^2\right].
    \label{eq:field-regression}
\end{equation}
The critic observes $\xx_\alpha$ and $\cc$, but not $\alpha$ separately, so it pools supervision across interpolation times into one field for the current pair of distributions. Implementation-specific normalization and loss scaling are in \appref{app:implementation}.
Regression averages pairwise targets into the population mismatch field
\begin{equation}
    \vv_{\mtheta}(\xx,\cc)
    =\E\!\left[\mX_{\mathrm{r}}-\mX_{\mathrm{p}}
    \mid \mX_\alpha=\xx,\mC=\cc\right],
    \label{eq:conditional-field}
\end{equation}
where capital letters denote the random variables of Eq.~\eqref{eq:corridor}. Squared-error regression recovers this conditional mean over unrestricted fields; over an idealized linear potential class, it recovers its conservative projection (\propref{prop:population-field}). The field is defined $\mu_{\mtheta}$-almost everywhere and changes with the generator.

Random pairing therefore has a distributional interpretation: when the conditional endpoint laws match, exchanging the endpoints and replacing $\alpha$ by $1-\alpha$ preserves the query and reverses the target. The conditional mean vanishes even though individual displacements remain noisy, so generated samples are not regressed directly toward randomly assigned real samples.

\subsection{Adversarial dual and distribution matching}
Because the field is a gradient, the fundamental theorem of calculus along Eq.~\eqref{eq:corridor} gives, for every pair,
\begin{equation}
    \int_0^1\left\langle\nabla_{\xx}u_{\mphi}(\xx_\alpha,\cc),\mDelta\right\rangle\,\dm\alpha
    =u_{\mphi}(\xx_{\mathrm{r}},\cc)-u_{\mphi}(\xx_{\mathrm{p}},\cc).
    \label{eq:line-integral}
\end{equation}
Expanding the square in Eq.~\eqref{eq:field-regression} and applying Eq.~\eqref{eq:line-integral} to the cross term yields
\begin{equation}
    \begin{aligned}
        \cL_{\mathrm{field}}(\mphi;\mtheta)
        &=\tfrac{1}{2}\E\|\mDelta\|_2^2-\cV(\mphi;\mtheta),\\[2pt]
        \cV(\mphi;\mtheta)
        &=\E[u_{\mphi}(\xx_{\mathrm{r}},\cc)-u_{\mphi}(\xx_{\mathrm{p}},\cc)]
        -\tfrac{1}{2}\E\|\nabla_{\xx}u_{\mphi}(\xx_\alpha,\cc)\|_2^2.
    \end{aligned}
    \label{eq:nextgan-value}
\end{equation}
The first term does not depend on $\mphi$, so minimizing the regression loss is \emph{exactly} maximizing the adversarial value $\cV$: a real--generated potential gap minus the field energy on the corridor. Its supremum over potentials defines the discrepancy
\begin{equation}
    \Dnext(P_{\mtheta},Q)
    =\sup_{u\in\cH}\left\{
    \E\big[u(\xx_{\mathrm{r}},\cc)-u(\xx_{\mathrm{p}},\cc)\big]
    -\frac{1}{2}\E\|\nabla_{\xx}u(\xx_\alpha,\cc)\|_2^2\right\}.
    \label{eq:distribution-objective}
\end{equation}
We train with both forms: the regression loss in Eq.~\eqref{eq:field-regression} and the adversarial critic loss $\cL_{\mathrm{D}}^{\mathrm{adv}}=-\cV$ (\appref{app:gan-code}). Their equivalence holds for a fixed generator after averaging over uniform interpolation times; their finite-minibatch updates need not coincide. The regression form supplies displacement targets, while the adversarial form exposes the distributional quantity measured at the optimum.

\noindent\begin{minipage}{\linewidth}
\begin{proposition}[Population mismatch field and distribution matching]
\label{prop:population-field}
Assume finite second moments. Let $\cH$ be a linear space of potentials measurable in $\cc$ and $C^1$ in $\xx$, with integrable endpoint values and $\nabla_{\xx}u\in L^2(\mu_{\mtheta})$. Let $\cG$ be the $L^2(\mu_{\mtheta})$-closure of $\{\nabla_{\xx}u:u\in\cH\}$.
\begin{enumerate}[label=(\alph*), nosep, leftmargin=16pt]
    \item Over all square-integrable vector fields, the minimizer of Eq.~\eqref{eq:field-regression} is $\vv_{\mtheta}$; over $\cG$, it is the orthogonal projection $\operatorname{Proj}_{\cG}\vv_{\mtheta}$.
    \item $\Dnext(P_{\mtheta},Q)=\frac12\|\operatorname{Proj}_{\cG}\vv_{\mtheta}\|_{L^2(\mu_{\mtheta})}^2\ge 0$.
    \item If $P_{\mtheta}(\cdot\mid\cc)=Q(\cdot\mid\cc)$ for almost every $\cc$, then $\vv_{\mtheta}=\zero$ and $\Dnext=0$. Conversely, if $\Dnext=0$ and $\cH$ determines the joint law of $(\mX,\mC)$---for example, if it contains every product $u(\xx,\cc)=a(\xx)b(\cc)$ with $a$ smooth and compactly supported and $b$ bounded and measurable---then the conditional feature distributions coincide for almost every $\cc$.
\end{enumerate}
\end{proposition}

The proof is in \appref{app:population-field}; finite potential networks restrict this population characterization to their parameterization.
\end{minipage}
\par

\begin{remark}[Negative-Sobolev form]
\label{rem:sobolev}
Fix $\cc$ and write $q$, $p_{\mtheta}$, and $\omega_{\mtheta}$ for the densities of $Q(\cdot\mid\cc)$, $P_{\mtheta}(\cdot\mid\cc)$, and $\mu_{\mtheta}(\cdot\mid\cc)$. Assume that the weighted Poisson problem admits a weak solution under suitable boundary or decay conditions. When $\cH$ is unrestricted, the optimal potential solves $-\nabla_{\xx}\!\cdot\!\big(\omega_{\mtheta}\nabla_{\xx}u^\star\big)=q-p_{\mtheta}$ weakly. Averaging over conditions gives $\Dnext=\frac12\,\E_{\cc}\|q-p_{\mtheta}\|^2_{\dot{H}^{-1}(\mu_{\mtheta}(\cdot\mid\cc))}$. This is a squared negative-Sobolev norm weighted by the corridor measure, related to the discrepancy underlying Sobolev GAN~\citep{mroueh2018sobolev}. The regression loss equals the negative adversarial value plus the displacement-dependent constant in Eq.~\eqref{eq:nextgan-value}.
\end{remark}

\subsection{Alternating critic and generator updates}
With the generator fixed, the critic learns the current displacement field. We then hold the critic fixed and update the generator by ascending the potential at generated endpoints:
\begin{equation}
    \cL_{\mathrm{gen}}(\mtheta;\mphi)
    =-\E\,u_{\mphi}(\xx_{\mathrm{p}},\cc),
    \qquad
    \xx_{\mathrm{p}}=\mf\!\left(\mg_{\mtheta}(\zz,\cc),\cc\right).
    \label{eq:generator}
\end{equation}
The chain rule pulls $\nabla_{\xx}u_{\mphi}(\xx_{\mathrm{p}},\cc)$ back through $\mf$ and $\mg_{\mtheta}$. After each generator change, fresh pairs update the critic to reflect the new mismatch. \algxref{alg:nextgan} summarizes the loop: the generator retains the standard potential-based GAN update, while displacement supervision shapes its critic. At inference, the critic and representers are discarded.

The update queries generated endpoints, whereas regression supervises interpolated features; \propref{prop:population-field} characterizes the field only $\mu_{\mtheta}$-almost everywhere. Since the corridor measure also changes with the generator, the update is not generally the full gradient of $\Dnext$ (\appref{app:geometry-variation}).

\begin{algorithm}[!ht]
\caption{\NextGAN training (one representer, unit weight). $\sg[\cdot]$ denotes stop gradient.}
\label{alg:nextgan}
\begin{algorithmic}[1]
\Require frozen representer $\mf$; critic steps per generator step $n_{\mathrm{critic}}$
\State Initialize generator $\mg_{\mtheta}$ and potential $u_{\mphi}$
\Repeat
    \State Sample $(\bar{\xx}_{\mathrm{r}},\cc)$, $\zz\sim p(\zz)$, and generate $\bar{\xx}_{\mathrm{p}}\gets\mg_{\mtheta}(\zz,\cc)$
    \State Compute $\xx_{\mathrm{r}}\gets\mf(\bar{\xx}_{\mathrm{r}},\cc)$ and $\xx_{\mathrm{p}}\gets\mf(\bar{\xx}_{\mathrm{p}},\cc)$
    \For{$k=1,\ldots,n_{\mathrm{critic}}$}
        \State Sample $\alpha\sim\mathcal{U}(0,1)$ and set $\xx_\alpha\gets\alpha\xx_{\mathrm{r}}+(1-\alpha)\sg[\xx_{\mathrm{p}}]$
        \State Update $\mphi$ by descending $\displaystyle
            \frac{1}{2}\left\|\nabla_{\xx}u_{\mphi}(\xx_\alpha,\cc)
            -\sg[\xx_{\mathrm{r}}-\xx_{\mathrm{p}}]\right\|_2^2$
    \EndFor
    \State Hold $\mphi$ fixed; update $\mtheta$ by descending $\displaystyle -u_{\mphi}(\xx_{\mathrm{p}},\cc)$
\Until{convergence}
\end{algorithmic}
\end{algorithm}

\subsection{Training dynamics and mode coverage}
\label{sec:direction}
To examine how this change in critic supervision affects the alternating updates in \algxref{alg:nextgan}, we track generated samples across a mixture of eight Gaussians. \NextGAN and a non-saturating GAN share the initialization, critic capacity, latent probes, and 30K-update budget. \figref{fig:toy8-comparison} reports samples from the same latent probes and mode imbalance, $8-N_{\mathrm{eff}}$, where $N_{\mathrm{eff}}$ is the effective mode count among samples captured by the target modes. Zero indicates balanced allocation across all eight modes. Full experimental details are in \appref{app:toy}.

\NextGAN reaches all eight modes and maintains a nearly balanced allocation, ending at $N_{\mathrm{eff}}=7.995$. GAN initially spreads across the ring but later concentrates on fewer modes, ending at $N_{\mathrm{eff}}=2.770$. The fractions of samples captured by the modes are similar (92.8\% and 93.3\%, respectively). Thus, in this controlled experiment, the learned guidance supports balanced mode coverage; the main difference is how generated mass is allocated across modes, despite similar captured fractions.

\begin{figure}[!htbp]
\centering
\input{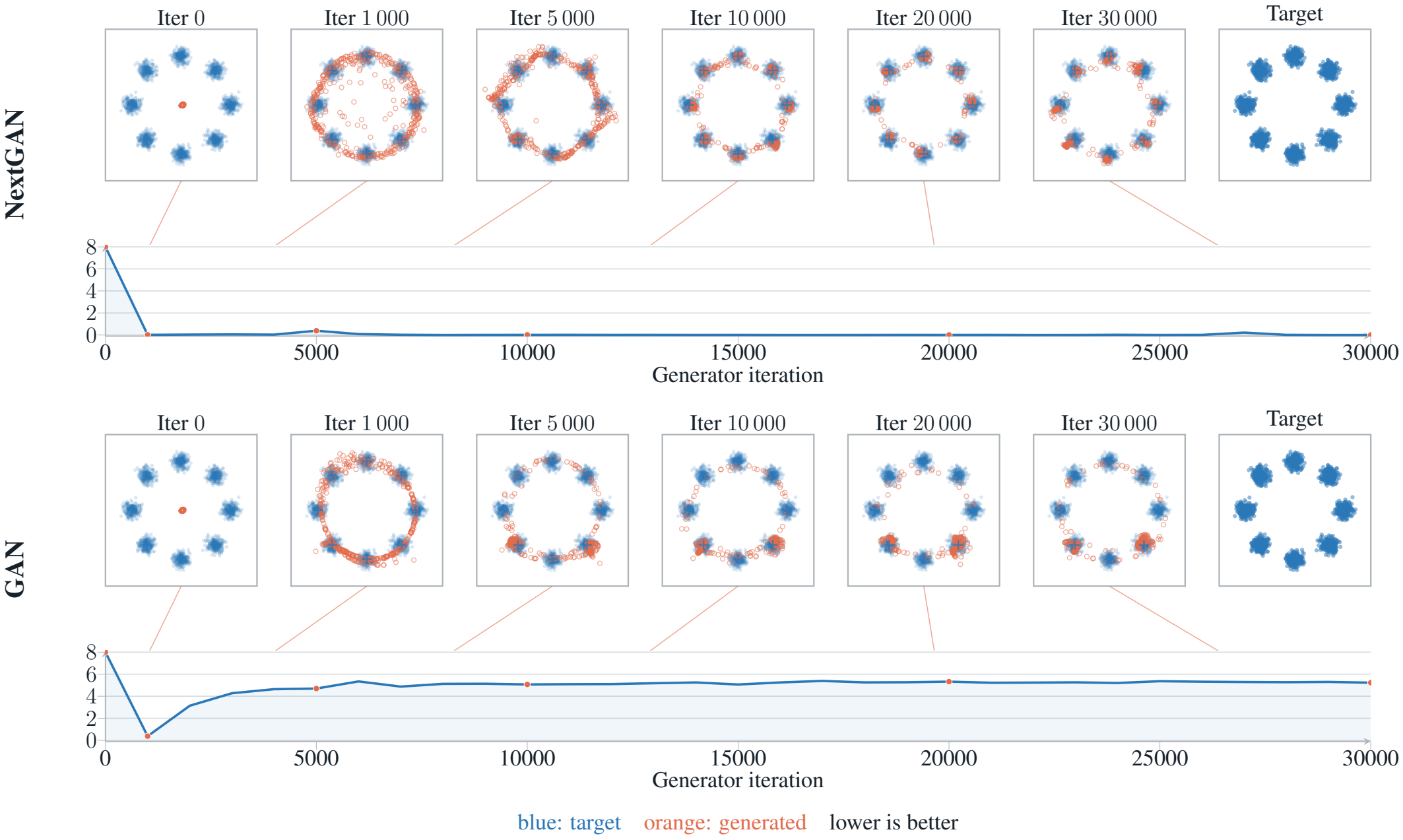}
\small\caption{\textbf{Mode coverage during training.} \emph{Rows:} \NextGAN and GAN. \emph{Columns:} samples from fixed latent probes at selected iterations and the target distribution. The curves below each row report mode imbalance, $8-N_{\mathrm{eff}}$; lower is better.}
\label{fig:toy8-comparison}
\end{figure}

\subsection{Practical implementation}
\label{sec:implementation}
\paragraph{Comparison spaces.}
We apply the same field-regression objective in frozen feature spaces. For images, we use frozen MAE-B/16, SigLIP2-B/16, and ResNet-18 encoders~\citep{he2022mae,tschn2025siglip2,he2016deep}. For video, we explored VideoMAE, V-JEPA2, InternVideo-Next, RVM, and DINOv3 as pretrained representers~\citep{tong2022videomae,assran2025vjepa2,wang2025internvideonext,zoran2025rvm,simeoni2025dinov3}. In our experiments, these features tended to favor static appearance over motion, and no single representer captured all aspects of video generation. We therefore use the frozen pretrained video backbone itself as the representer. The generator uses LoRA~\citep{hu2022lora}; disabling the adapters provides frozen, text-conditioned features of clean video latents at selected layers.

\paragraph{Potential heads and optimization.}
Each image encoder or video layer feeds a separate four-block transformer head that outputs a conditional scalar potential. We sum the head losses and backpropagate through the frozen representers to the generator. We alternate $n_{\mathrm{critic}}$ critic updates with one generator update and maintain a generator EMA (\algxref{alg:nextgan}).

\paragraph{Backward simulation.}
For video, following the backward simulation strategy of rCM~\citep{rcm2025}, training uses randomly sampled rollout lengths. Starting from pure noise, we alternate denoising to a clean estimate using the Wan velocity prediction and forward noising with fresh noise, keeping noise levels non-increasing. Here $\mg_{\mtheta}$ denotes the complete sampling procedure, with $\zz$ collecting its random inputs. We backpropagate the generator loss only through the final denoising step; earlier rollout steps are detached, yielding a truncated gradient through this procedure.

\FloatBarrier
\section{Experiments}
\label{sec:experiments}
Following the controlled analysis in \secref{sec:direction}, we evaluate \NextGAN on pixel- and latent-space image generators and on video generators with up to 14B parameters.

\paragraph{Protocol.}
We evaluate gradient regression and its equivalent adversarial form, both with the standard GAN generator update. Training requires no online teacher scores or trajectories. \appref{app:implementation} details the architectures, checkpoint initialization, evaluation settings, and training compute.

\subsection{ImageNet-256: one-step generation}
\label{sec:imagenet}
We apply the same objective to class-conditional ImageNet-256 generation~\citep{deng2009imagenet} in both latent and pixel space. \tabref{tab:imagenet} reports quality and sampling cost, including evaluations used for classifier-free guidance~\citep{ho2022classifierfree}. Gradient regression achieves 1.54 FID with the latent DiT-XL/1 and 2.87 with the pixel-space JiT-B, each in one network evaluation without guidance. The equivalent adversarial form achieves 1.36 FID with DiT-XL/1. These results show that the training objective works with both generator types. \figref{fig:qualitative-overview} shows samples from the latent model.

\begin{table}[t]
\caption{Class-conditional ImageNet-256 results. Baselines are quoted from their original papers and may use different evaluators. \NextGAN Params denote nominal backbone sizes (implementation counts: App.~\ref{app:implementation}). NFE counts sequential generator evaluations, including guidance and autoregressive inner steps where applicable.}
\label{tab:imagenet}
\centering
\begingroup\scriptsize
\setlength{\tabcolsep}{2.6pt}
\begin{tabularx}{\linewidth}{@{}p{0.25\linewidth}p{0.07\linewidth}p{0.17\linewidth}p{0.11\linewidth}p{0.10\linewidth}YY@{}}
\toprule
Method & Space & Backbone & Params & NFE & FID $\downarrow$ & IS $\uparrow$ \\
\midrule
\multicolumn{7}{@{}l}{\emph{Consistency / flow-map distillation}}\\[-1pt]
Shortcut~\citep{frans2025shortcut} & latent & DiT-XL/2 & 676M & 1 & 10.60 & -- \\
PaGoDA~\citep{kim2024pagoda} & pixel & ADM & 301M & 1 & 1.56 & 259.61 \\
\midrule
\multicolumn{7}{@{}l}{\emph{Diffusion / flow matching}}\\[-1pt]
MeanFlow~\citep{geng2025meanflows} & latent & XL/2 & 676M & 1 & 3.43 & -- \\
ADM~\citep{dhariwal2021diffusion} & pixel & ADM-G + ADM-U & 866M & $250{+}250$ & 3.94 & 215.8 \\
VDM++~\citep{kingma2023vdmpp} & pixel & U-ViT-L/2 & 2B & $512{\times}2$ & 2.12 & 267.7 \\
DiT~\citep{peebles2023dit} & latent & DiT-XL/2 & 675M & $250{\times}2$ & 2.27 & 278.2 \\
SiT~\citep{ma2024sit} & latent & SiT-XL/2 & 675M & $250{\times}2$ & 2.06 & 277.5 \\
JiT~\citep{li2025jit} & pixel & JiT-H/16 & 953M & 200 & 1.86 & 303.4 \\
\midrule
\multicolumn{7}{@{}l}{\emph{Autoregressive models}}\\[-1pt]
VAR~\citep{tian2024var} & discrete & VAR-d30 & 2B & $10{\times}2$ & 1.92 & 323.1 \\
BAR~\citep{yu2026bar} & discrete & BAR-L & 1.1B & 2,048 & 0.99 & 296.9 \\
MAR~\citep{li2024mar} & latent & MAR-H & 943M & 51,200 & 1.55 & 303.7 \\
\midrule
\multicolumn{7}{@{}l}{\emph{GAN-based one-step generators}}\\[-1pt]
BigGAN-deep~\citep{brock2019biggan} & pixel & BigGAN-deep & 112M & 1 & 6.95 & 171.4 \\
GigaGAN~\citep{kang2023gigagan} & pixel & GigaGAN & 569M & 1 & 3.45 & 225.5 \\
StyleGAN-XL~\citep{sauer2022styleganxl} & pixel & StyleGAN-XL & 166M & 1 & 2.30 & 265.1 \\
\midrule
\multicolumn{7}{@{}l}{\emph{Direct distribution dynamics and distributional losses}}\\[-1pt]
Drift~\citep{deng2026drifting} & pixel & DiT-L/16 & 464M & 1 & 1.61 & 307.5 \\
FD-loss~\citep{yang2026fdloss} & pixel & JiT-H/16 & 953M & 1 & 0.75 & 313.0 \\
TBSM~\citep{sun2026threebody} & latent & DiT-XL/2 & 675M & 1 & 1.63 & 243.0 \\
\midrule
\multicolumn{7}{@{}l}{\emph{\NextGAN (no online teacher queries)}}\\[-1pt]
\rowcolor{nextganpale}
\NextGAN & pixel & JiT-B & 131M & 1 & 2.87 & 207.1 \\
\rowcolor{nextganpale}
\NextGAN & latent & DiT-XL/1 & 675M & 1 & 1.54 & 243.5 \\
\rowcolor{nextganpale}
\NextGAN (adversarial form) & latent & DiT-XL/1 & 675M & 1 & 1.36 & 241.7 \\
\bottomrule
\end{tabularx}
\endgroup
\end{table}
\FloatBarrier

\appref{app:ablations} compares training objectives and frozen representers on JiT-B, with the training curves and feature-space ablations in \figref{fig:mechanism-signal}.

\subsection{Text-to-video generation}
\label{sec:video}
We fine-tune Wan2.1 text-to-video generators~\citep{wan2025wan} at 1.3B and 14B parameters using LoRA~\citep{hu2022lora} on the publicly released Wan\_datasets from rCM~\citep{rcm2025}. We use global batch sizes of 32 and 16 for the 1.3B and 14B models, respectively. With its adapters disabled, the frozen pretrained Wan2.1 backbone serves as the representer and supplies intermediate features to the potential heads. These heads apply joint attention to text and video tokens. The field-regression objective is otherwise unchanged.

At each model scale, we evaluate one-, two-, and four-step sampling using the backward simulation procedure described in \secref{sec:implementation}. Evaluation details are provided in \appref{app:implementation}.

\begin{table}[t]
\caption{Text-to-video generation on VBench at $832\times480\times81$. Results are grouped by model size and NFE. ``Online teacher'' indicates whether training requires online teacher scores or trajectories. Higher is better for all three scores.}
\label{tab:vbench}
\centering\scriptsize
\renewcommand{\arraystretch}{0.9}
\setlength{\tabcolsep}{2.6pt}
\begin{tabularx}{\linewidth}{@{}p{0.30\linewidth}>{\centering\arraybackslash}p{0.14\linewidth}>{\centering\arraybackslash}p{0.08\linewidth}YYY@{}}
\toprule
Model & Online teacher & NFE & \multicolumn{3}{c}{Evaluation Scores $\uparrow$} \\
\cmidrule(l){4-6}
 & & & Quality & Semantic & Total \\
\midrule
\multicolumn{6}{@{}l}{\emph{1.3B parameters}}\\[-1pt]
Wan2.1 T2V~\citep{wan2025wan} & -- & $50\times2$ & 82.91 & 80.22 & 82.37 \\
rCM~\citep{rcm2025} & yes & 1 & 81.76 & \best{76.34} & 80.68 \\
AnyFlow~\citep{anyflow2026} & yes & 1 & 78.74 & 75.63 & 78.12 \\
Wan2.1 $+$ DMD2~\citep{yin2024improved} & yes & 1 & 80.30 & 71.85 & 78.61 \\
\rowcolor{nextganpale}
\NextGAN & no & 1 & \best{82.15} & 75.41 & \best{80.80} \\
rCM~\citep{rcm2025} & yes & 2 & 82.61 & 78.99 & 81.88 \\
AnyFlow~\citep{anyflow2026} & yes & 2 & 83.08 & \best{80.95} & 82.66 \\
Wan2.1 $+$ DMD2~\citep{yin2024improved} & yes & 2 & \best{84.08} & 77.56 & 82.78 \\
\rowcolor{nextganpale}
\NextGAN & no & 2 & 83.98 & 79.73 & \best{83.13} \\
rCM~\citep{rcm2025} & yes & 4 & 82.61 & 78.55 & 81.79 \\
AnyFlow~\citep{anyflow2026} & yes & 4 & 84.01 & \best{81.67} & \best{83.54} \\
Wan2.1 $+$ DMD2~\citep{yin2024improved} & yes & 4 & \best{84.13} & 76.45 & 82.59 \\
\rowcolor{nextganpale}
\NextGAN & no & 4 & 83.79 & 78.44 & 82.72 \\
\midrule
\multicolumn{6}{@{}l}{\emph{14B parameters}}\\[-1pt]
Wan2.1 T2V~\citep{wan2025wan} & -- & $50\times2$ & 84.26 & 80.92 & 83.58 \\
rCM~\citep{rcm2025} & yes & 1 & 80.04 & 74.33 & 78.90 \\
AnyFlow~\citep{anyflow2026} & yes & 1 & \best{81.36} & \best{80.71} & \best{81.23} \\
\rowcolor{nextganpale}
\NextGAN & no & 1 & 79.78 & 79.71 & 79.77 \\
rCM~\citep{rcm2025} & yes & 2 & 83.72 & 80.58 & 83.09 \\
AnyFlow~\citep{anyflow2026} & yes & 2 & \best{84.12} & \best{82.97} & \best{83.89} \\
\rowcolor{nextganpale}
\NextGAN & no & 2 & 82.82 & 82.41 & 82.73 \\
rCM~\citep{rcm2025} & yes & 4 & 83.66 & 81.22 & 83.17 \\
AnyFlow~\citep{anyflow2026} & yes & 4 & \best{84.36} & \best{83.07} & \best{84.10} \\
\rowcolor{nextganpale}
\NextGAN & no & 4 & 82.24 & 81.14 & 82.02 \\
\bottomrule
\end{tabularx}
\end{table}

\tabref{tab:vbench} reports self-evaluated VBench results~\citep{huang2024vbench}; the CausVid DMD2 entries use the publicly released bidirectional checkpoint. At 1.3B parameters, a single model reaches total scores of 80.80, 83.13, and 82.72 with one, two, and four evaluations, respectively. The 14B model reaches total scores of 79.77, 82.73, and 82.02 at one, two, and four steps. AnyFlow retains the highest total score among the compared 14B methods at each budget; its two-step semantic score (82.97) also exceeds \NextGAN (82.41).

\paragraph{Comparison with GAN, WGAN, and TBSM (GAN).}
\figref{fig:video-objectives} compares four-step video generation across GAN, WGAN, TBSM (GAN), and \NextGAN. At 1.3B parameters, \NextGAN attains a VBench total score of 82.72, compared with 81.38 for GAN, 60.71 for WGAN, and 73.07 for TBSM (GAN). The selected qualitative examples show clearer object identity and scene structure with \NextGAN.

\begin{figure}[H]
\centering
\begingroup
\noindent\begin{minipage}[t]{0.40\linewidth}
\vspace{0pt}
\begingroup
\definecolor{vbgan}{HTML}{6486AD}
\definecolor{vbwgan}{HTML}{B290BD}
\definecolor{vbtbsm}{HTML}{DEA34C}
\definecolor{vbnext}{HTML}{B54445}
\centering
\scriptsize
\setlength{\tabcolsep}{3pt}
\newcommand{\vbswatch}[1]{\textcolor{#1}{\rule{7pt}{5pt}}}
\begin{tabular}{@{}ll@{}}
\vbswatch{vbgan} GAN & \vbswatch{vbwgan} WGAN \\
\vbswatch{vbtbsm} TBSM (GAN) & \vbswatch{vbnext} \textcolor{vbnext}{\textbf{\NextGAN}}
\end{tabular}\par\vspace{4pt}
\begin{tikzpicture}
\begin{axis}[
  width=\dimexpr\linewidth-23pt\relax,height=3.05cm,scale only axis,
  xmin=-0.6,xmax=13.6,ymin=0,ymax=105,
  xtick={1.5,6.5,11.5},xticklabels={Quality,Semantic,Total},
  ytick={0,25,50,75,100},
  tick label style={font=\tiny},
  axis lines*=left,axis line style={gray!55},
  tick style={gray!55},tick align=outside,
  ymajorgrids=true,grid style={gray!18,line width=0.3pt},
  ybar,bar width=7.2pt,bar shift=0pt,
  nodes near coords,nodes near coords style={font=\fontsize{4.5}{5}\selectfont,inner sep=1pt,/pgf/number format/fixed,/pgf/number format/precision=2,/pgf/number format/fixed zerofill},
  clip=false,
]
\addplot[fill=vbgan,draw=none] coordinates {(0,82.29) (5,77.78) (10,81.38)};
\addplot[fill=vbwgan,draw=none] coordinates {(1,70.09) (6,23.17) (11,60.71)};
\addplot[fill=vbtbsm,draw=none] coordinates {(2,80.19) (7,44.57) (12,73.07)};
\addplot[fill=vbnext,draw=none,nodes near coords style={font=\bfseries\boldmath\fontsize{4.5}{5}\selectfont}] coordinates {(3,83.79) (8,78.44) (13,82.72)};
\end{axis}
\end{tikzpicture}
\endgroup

\end{minipage}\hfill
\begin{minipage}[t]{0.58\linewidth}
\vspace{0pt}
\setlength{\tabcolsep}{1.2pt}
\renewcommand{\arraystretch}{1.0}
\scriptsize
\newcommand{\vbenchframe}[2]{\includegraphics[width=\linewidth]{figures/video_objectives/#1/#2_selected.jpg}}
\begin{tabularx}{\linewidth}{@{}*{4}{>{\centering\arraybackslash}X}@{}}
\textbf{GAN} & \textbf{WGAN} & \textbf{TBSM (GAN)} & \textcolor[HTML]{B54445}{\textbf{\NextGAN}} \\[2pt]
\vbenchframe{01_yoda_guitar}{GAN} & \vbenchframe{01_yoda_guitar}{WGAN} & \vbenchframe{01_yoda_guitar}{TBSM} & \vbenchframe{01_yoda_guitar}{NextGAN} \\[2pt]
\vbenchframe{02_steam_train}{GAN} & \vbenchframe{02_steam_train}{WGAN} & \vbenchframe{02_steam_train}{TBSM} & \vbenchframe{02_steam_train}{NextGAN} \\[2pt]
\vbenchframe{04_stop_sign}{GAN} & \vbenchframe{04_stop_sign}{WGAN} & \vbenchframe{04_stop_sign}{TBSM} & \vbenchframe{04_stop_sign}{NextGAN} \\
\end{tabularx}
\end{minipage}
\endgroup

\small\caption{\textbf{Video generation with GAN, WGAN, TBSM (GAN), and \NextGAN.} \emph{Left:} \NextGAN achieves the highest quality, semantic, and total VBench scores (best in bold). \emph{Right:} \NextGAN shows clearer subject identity, more coherent object structure, and more accurate text generation.}
\label{fig:video-objectives}
\end{figure}
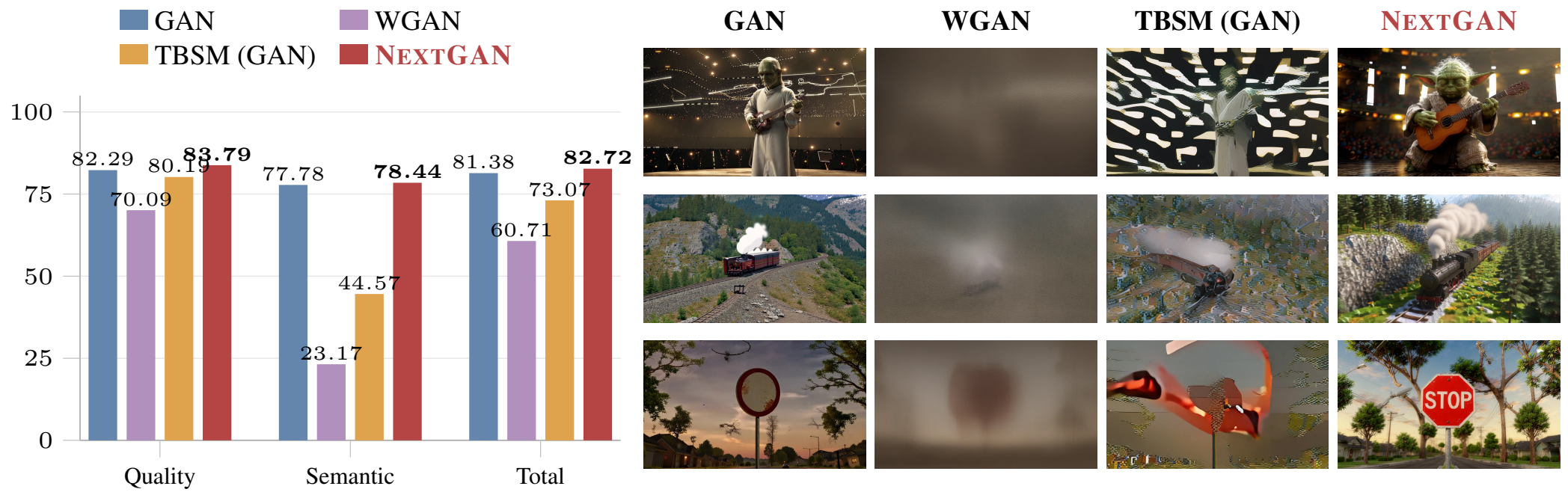

\flushbottom
\setlength{\intextsep}{4pt plus 1fil}
\paragraph{Qualitative comparison.}
To complement the aggregate VBench scores, \figref{fig:video-qualitative} compares single frames from rCM, AnyFlow, and \NextGAN under matched text prompts, with frames manually selected at matching indices across methods. In these examples, \NextGAN produces more realistic visual appearances than rCM and AnyFlow. Matching prompts and frame indices makes the visual comparison consistent across methods. These still-frame examples complement the aggregate evaluation: they illustrate visual differences, while temporal behavior must be assessed from the videos and the corresponding video-quality metrics.

\begin{figure}[H]
\centering
\begingroup
\setlength{\tabcolsep}{0.75pt}
\renewcommand{\arraystretch}{1.05}
\scriptsize
\font\videolabelfont=phvb at 8pt
\newcommand{\videocaseframe}[1]{\raisebox{-0.5\height}{\includegraphics[width=\linewidth]{figures/video_qualitative_frames/#1}}}
\begin{tabularx}{\linewidth}{@{}*{6}{>{\centering\arraybackslash}X}@{}}
\multicolumn{6}{@{}l@{}}{\videolabelfont\color{videobaselinelabel}rCM} \\[1pt]
\videocaseframe{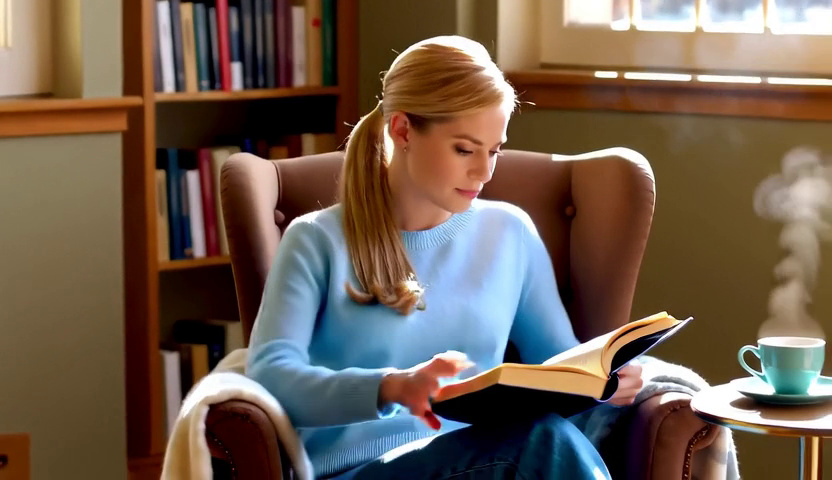} & \videocaseframe{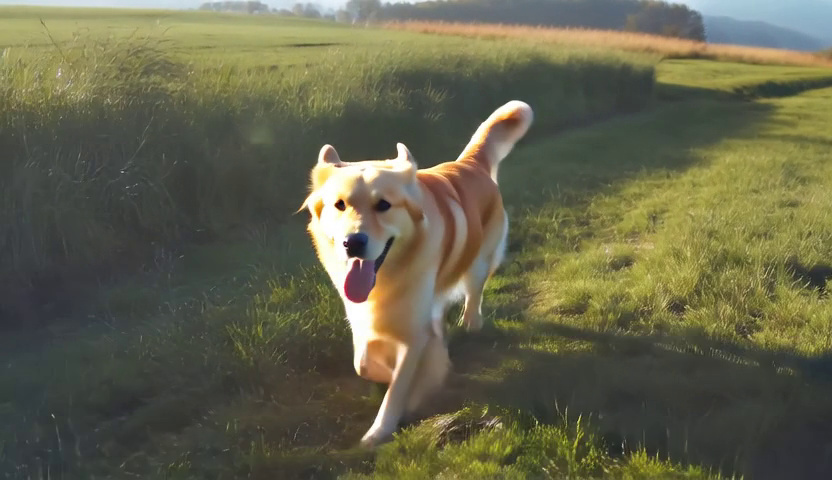} & \videocaseframe{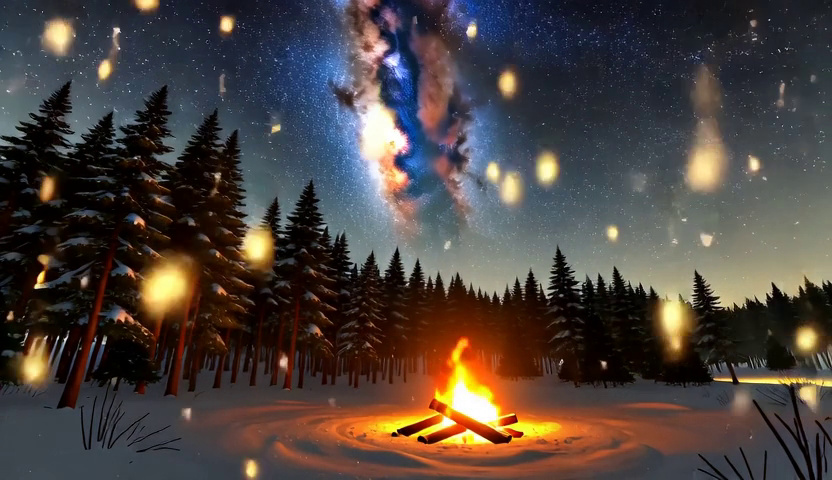} & \videocaseframe{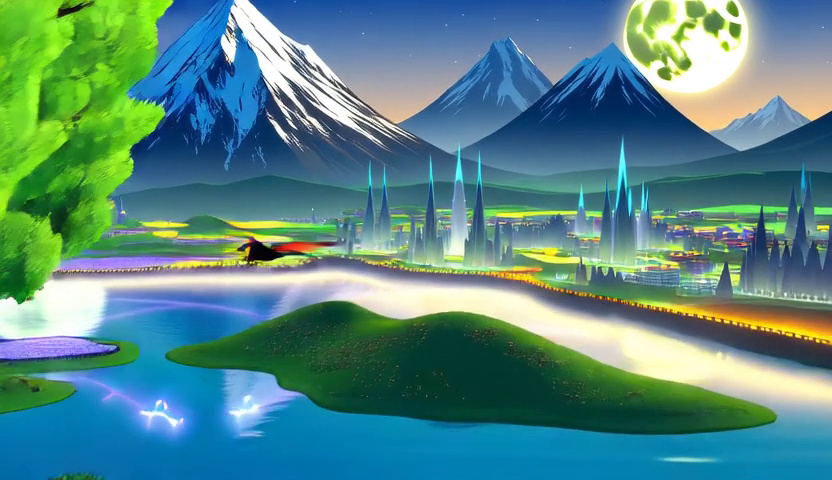} & \videocaseframe{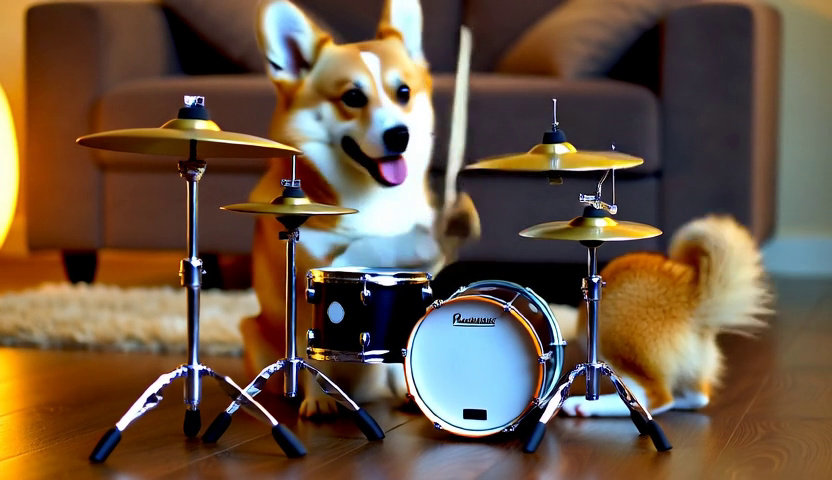} & \videocaseframe{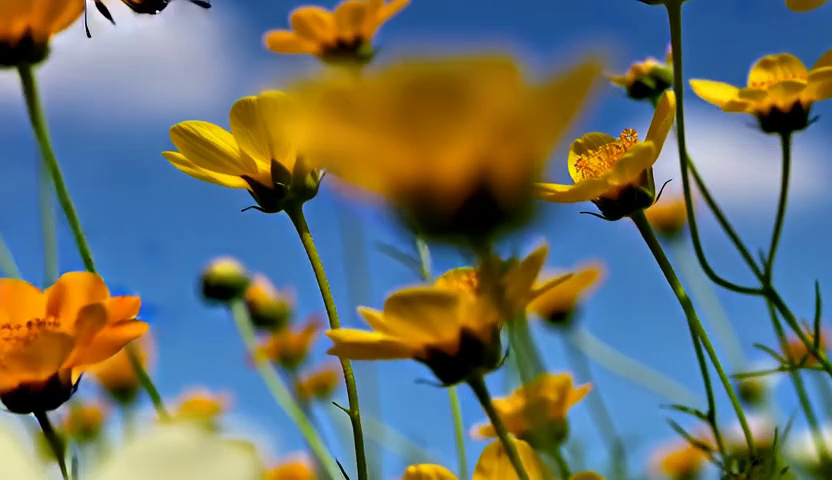} \\
\noalign{\vskip 1pt}
\multicolumn{6}{@{}l@{}}{\videolabelfont\color{videobaselinelabel}AnyFlow} \\[1pt]
\videocaseframe{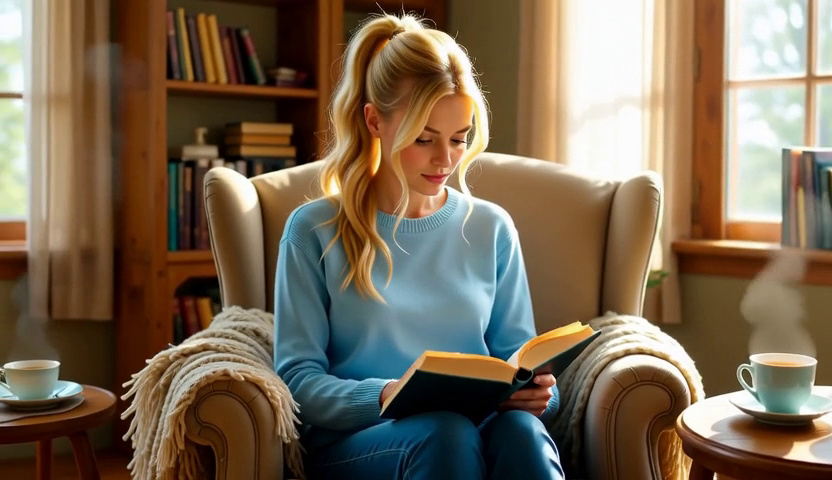} & \videocaseframe{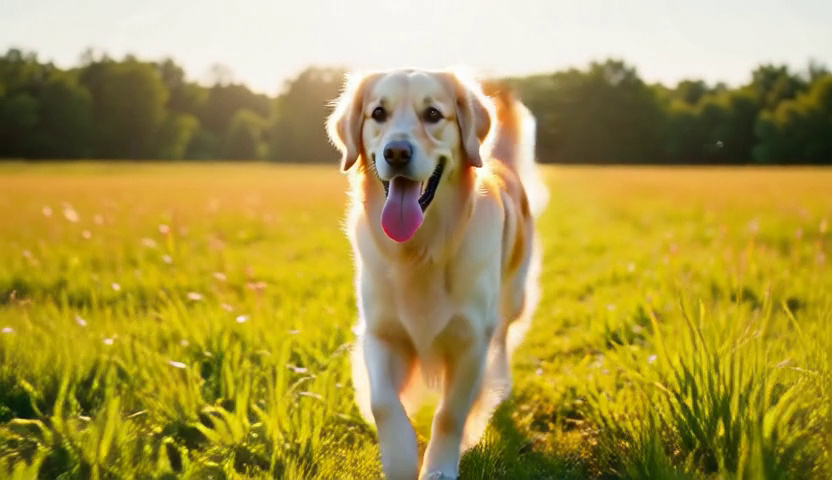} & \videocaseframe{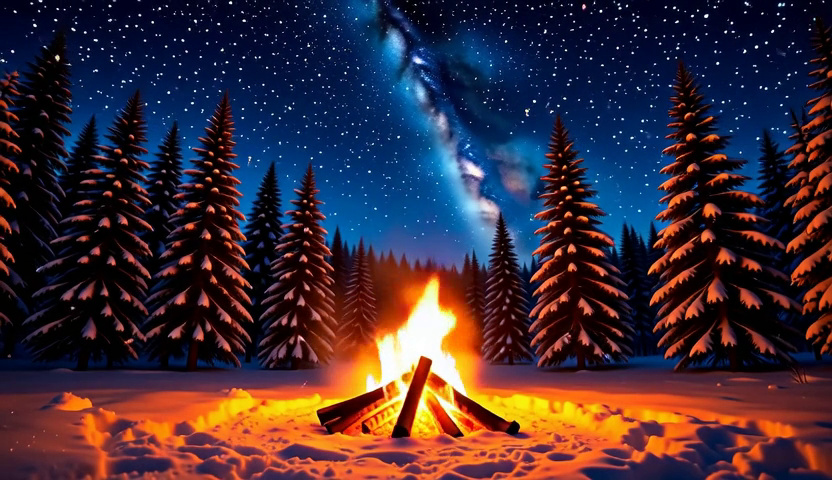} & \videocaseframe{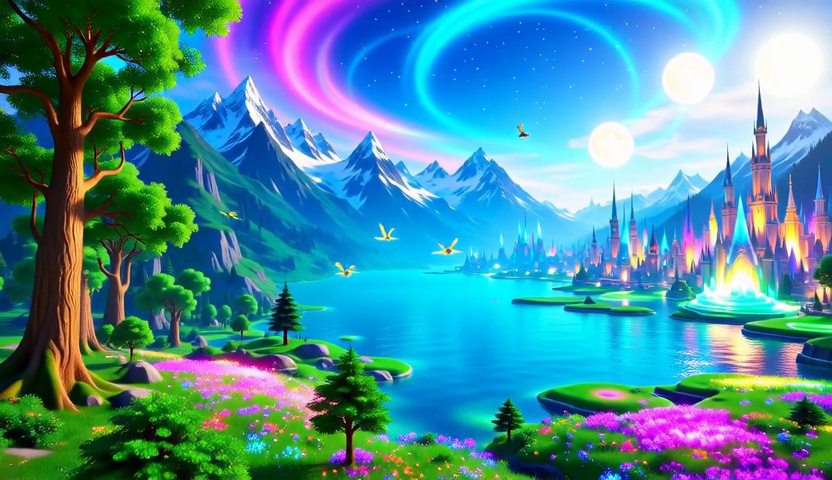} & \videocaseframe{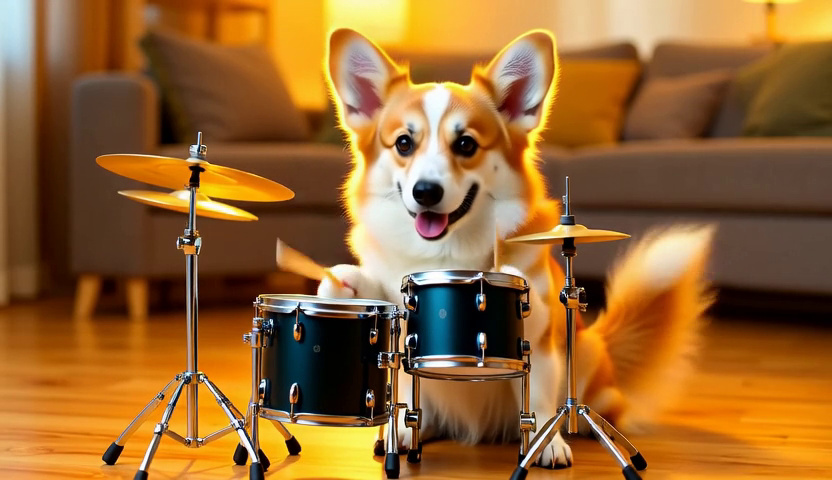} & \videocaseframe{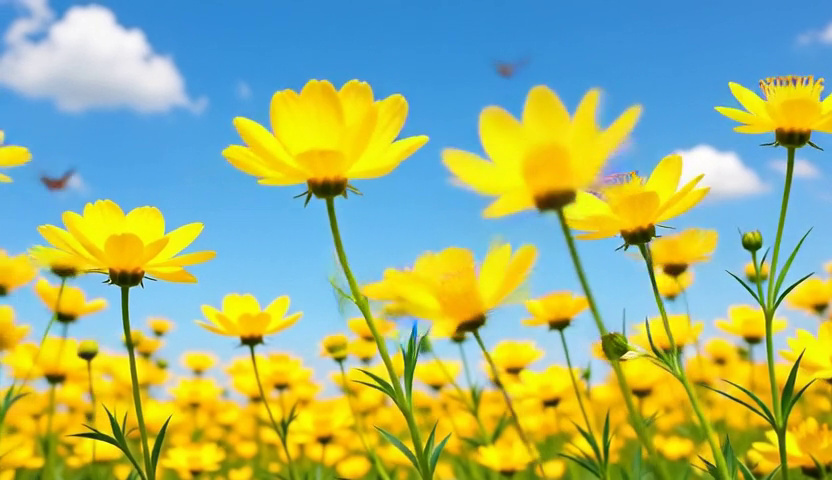} \\
\noalign{\vskip 1pt}
\multicolumn{6}{@{}l@{}}{\videolabelfont\color{videonextganlabel}NextGAN} \\[1pt]
\videocaseframe{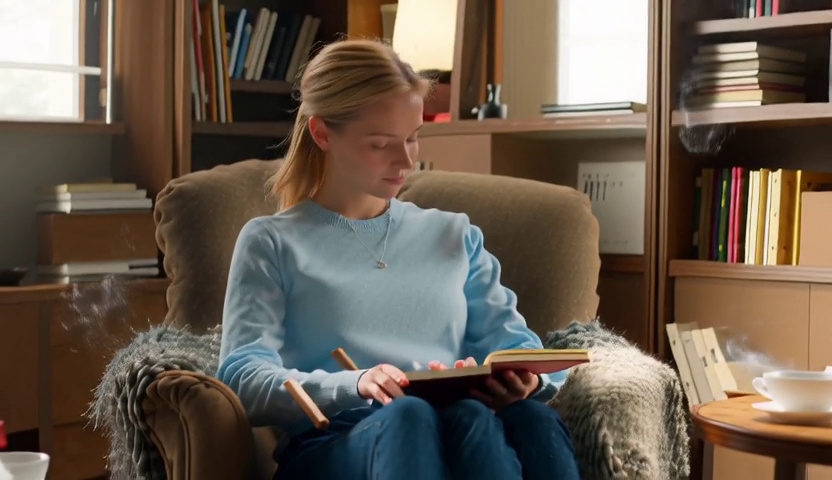} & \videocaseframe{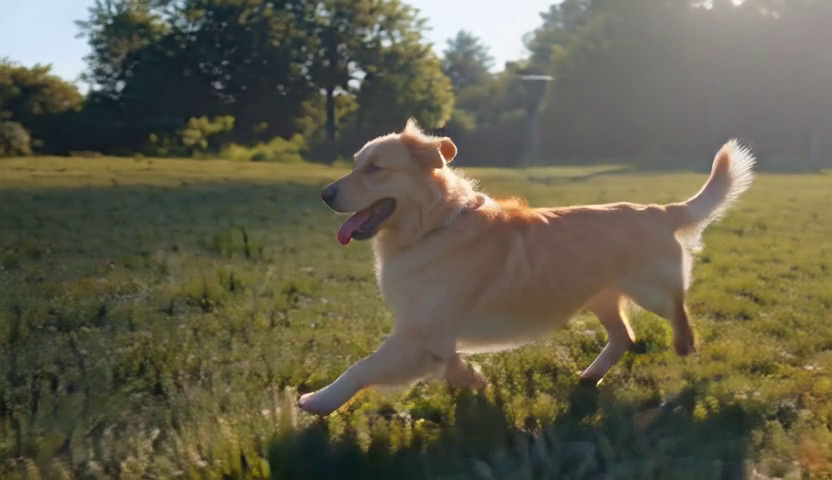} & \videocaseframe{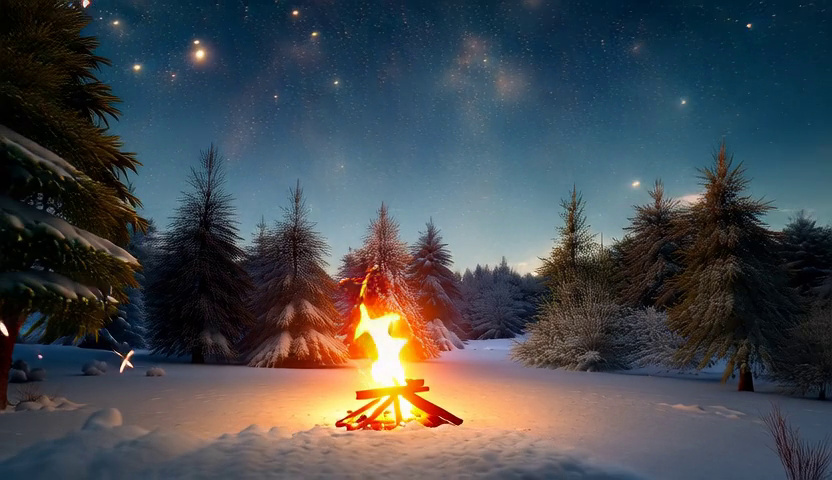} & \videocaseframe{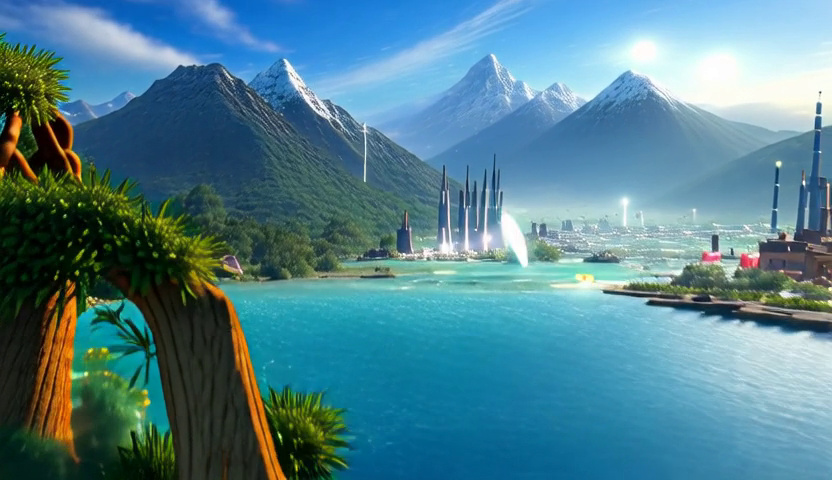} & \videocaseframe{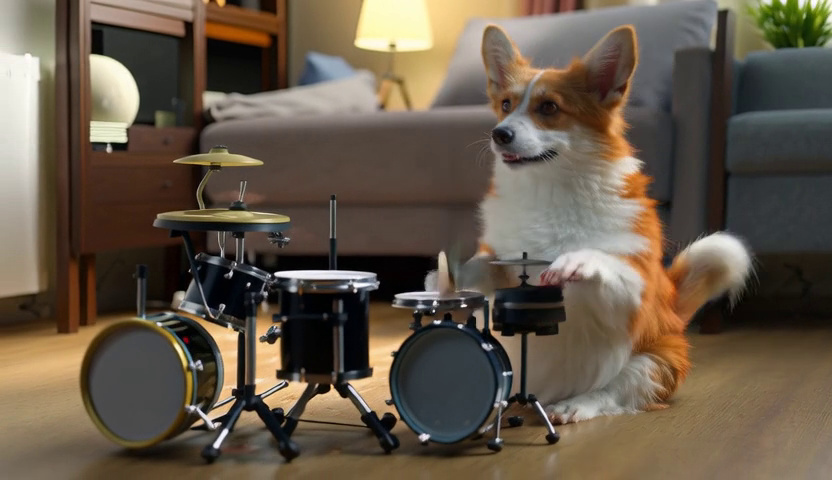} & \videocaseframe{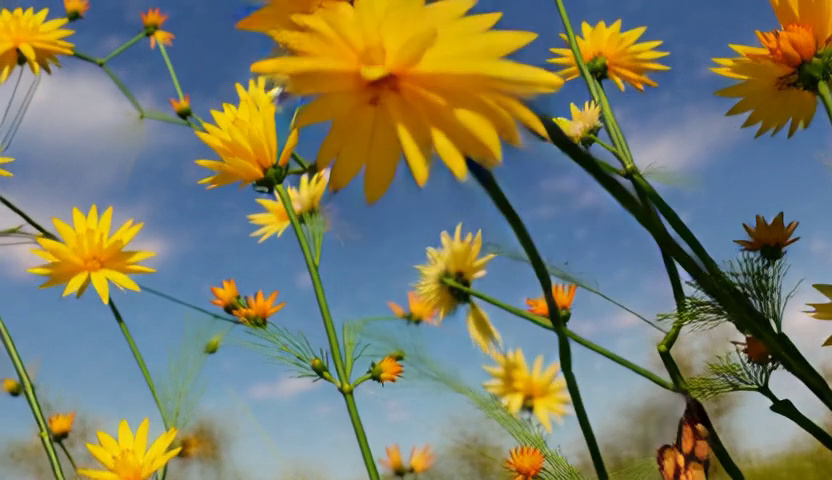} \\
\end{tabularx}
\endgroup
\small\caption{Six manually selected prompts (columns) with manually selected frames at matching indices across methods. Rows: rCM, AnyFlow, and \NextGAN (all 1.3B).}
\label{fig:video-qualitative}
\end{figure}

\paragraph{Sample diversity.}
\figref{fig:video-diversity} compares four-step samples from Wan2.1$+$DMD2~\citep{yin2024improved} and \NextGAN under matched prompts and per-prompt random seeds, using the same frame index for each corresponding pair of videos. Each row is one prompt and each column is a different seed; the left panel shows DMD2 and the right panel shows \NextGAN. Across these prompts, \NextGAN shows greater sample diversity than DMD2, with more varied layouts, colors, and subject configurations. Unlike the cross-method examples in \figref{fig:video-qualitative}, this comparison emphasizes variation across seeds for a fixed prompt. The two figures therefore provide complementary views of visual quality and within-prompt diversity.

\begin{figure}[H]
\centering
\input{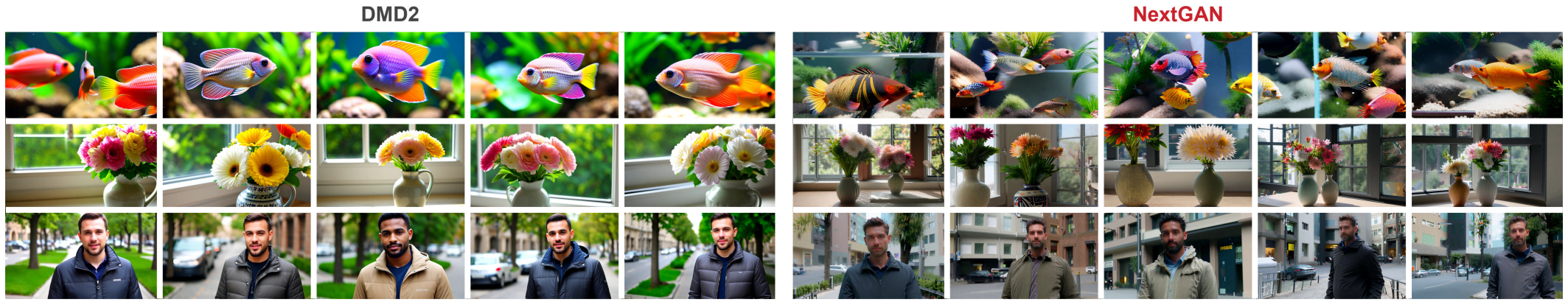}
\small\caption{Intra-prompt sample diversity under four-step sampling (Wan2.1-1.3B). Left: Wan2.1$+$DMD2; right: \NextGAN. Rows show fish, flowers, and man; each panel shows five seeds per prompt, with matched seed indices across methods. Corresponding videos from both methods are shown at the same frame index.}
\label{fig:video-diversity}
\end{figure}

\FloatBarrier
\section{Conclusion}
\NextGAN preserves the GAN interface while directly supervising the critic gradient consumed by the generator. Regression over random real--fake pairs recovers a population mismatch field, and a conservative parameterization gives the population critic objective an exact adversarial dual over a negative-Sobolev discrepancy. The same objective fine-tunes pixel- and latent-space image generators and video generators with up to 14B parameters, supporting multiple inference budgets. The matching guarantee applies in the chosen representation spaces under the stated assumptions. Training large generators from scratch remains future work.

\par\pagebreak
\endgroup
\linespread{1}\normalsize
\bibliography{sources/references}

\begin{thebibliography}{49}
\providecommand{\natexlab}[1]{#1}
\providecommand{\url}[1]{\texttt{#1}}
\expandafter\ifx\csname urlstyle\endcsname\relax
  \providecommand{\doi}[1]{doi: #1}\else
  \providecommand{\doi}{doi: \begingroup \urlstyle{rm}\Url}\fi

\bibitem[Arjovsky et~al.(2017)Arjovsky, Chintala, and
  Bottou]{arjovsky2017wasserstein}
Martin Arjovsky, Soumith Chintala, and L{\'e}on Bottou.
\newblock Wasserstein generative adversarial networks.
\newblock In \emph{Proceedings of the 34th International Conference on Machine
  Learning}, volume~70 of \emph{Proceedings of Machine Learning Research}, pp.\
   214--223, 2017.
\newblock URL \url{https://proceedings.mlr.press/v70/arjovsky17a.html}.

\bibitem[Assran et~al.(2025)Assran, Bardes, Fan, Garrido, Howes, Komeili,
  Muckley, Rizvi, Roberts, Sinha, Zholus, Arnaud, Gejji, Martin, Hogan, Dugas,
  Bojanowski, Khalidov, Labatut, Massa, Szafraniec, Krishnakumar, Li, Ma,
  Chandar, Meier, LeCun, Rabbat, and Ballas]{assran2025vjepa2}
Mido Assran, Adrien Bardes, David Fan, Quentin Garrido, Russell Howes, Mojtaba
  Komeili, Matthew Muckley, Ammar Rizvi, Claire Roberts, Koustuv Sinha, Artem
  Zholus, Sergio Arnaud, Abha Gejji, Ada Martin, Francois~Robert Hogan, Daniel
  Dugas, Piotr Bojanowski, Vasil Khalidov, Patrick Labatut, Francisco Massa,
  Marc Szafraniec, Kapil Krishnakumar, Yong Li, Xiaodong Ma, Sarath Chandar,
  Franziska Meier, Yann LeCun, Michael Rabbat, and Nicolas Ballas.
\newblock {V-JEPA 2}: Self-supervised video models enable understanding,
  prediction and planning.
\newblock \emph{arXiv preprint arXiv:2506.09985}, 2025.

\bibitem[Benamou \& Brenier(2000)Benamou and Brenier]{benamou2000computational}
Jean-David Benamou and Yann Brenier.
\newblock A computational fluid mechanics solution to the {Monge--Kantorovich}
  mass transfer problem.
\newblock \emph{Numerische Mathematik}, 84\penalty0 (3):\penalty0 375--393,
  2000.
\newblock \doi{10.1007/s002110050002}.

\bibitem[Brock et~al.(2019)Brock, Donahue, and Simonyan]{brock2019biggan}
Andrew Brock, Jeff Donahue, and Karen Simonyan.
\newblock Large scale {GAN} training for high fidelity natural image synthesis.
\newblock In \emph{International Conference on Learning Representations}, 2019.
\newblock URL \url{https://openreview.net/forum?id=B1xsqj09Fm}.

\bibitem[Deng et~al.(2009)Deng, Dong, Socher, Li, Li, and
  Fei-Fei]{deng2009imagenet}
Jia Deng, Wei Dong, Richard Socher, Li-Jia Li, Kai Li, and Li~Fei-Fei.
\newblock {ImageNet}: A large-scale hierarchical image database.
\newblock In \emph{IEEE Conference on Computer Vision and Pattern Recognition},
  pp.\  248--255, 2009.
\newblock \doi{10.1109/CVPR.2009.5206848}.

\bibitem[Deng et~al.(2026)Deng, Li, Li, Du, and He]{deng2026drifting}
Mingyang Deng, He~Li, Tianhong Li, Yilun Du, and Kaiming He.
\newblock Generative modeling via drifting.
\newblock \emph{arXiv preprint arXiv:2602.04770}, 2026.
\newblock URL \url{https://arxiv.org/abs/2602.04770}.

\bibitem[Dhariwal \& Nichol(2021)Dhariwal and Nichol]{dhariwal2021diffusion}
Prafulla Dhariwal and Alex Nichol.
\newblock Diffusion models beat {GANs} on image synthesis.
\newblock In \emph{Advances in Neural Information Processing Systems},
  volume~34, pp.\  8780--8794, 2021.

\bibitem[Frans et~al.(2025)Frans, Hafner, Levine, and
  Abbeel]{frans2025shortcut}
Kevin Frans, Danijar Hafner, Sergey Levine, and Pieter Abbeel.
\newblock One step diffusion via shortcut models.
\newblock In \emph{International Conference on Learning Representations}, 2025.
\newblock URL \url{https://arxiv.org/abs/2410.12557}.

\bibitem[Geng et~al.(2025)Geng, Deng, Bai, Kolter, and He]{geng2025meanflows}
Zhengyang Geng, Mingyang Deng, Xingjian Bai, J.~Zico Kolter, and Kaiming He.
\newblock Mean flows for one-step generative modeling.
\newblock \emph{arXiv preprint arXiv:2505.13447}, 2025.
\newblock URL \url{https://arxiv.org/abs/2505.13447}.

\bibitem[Goodfellow et~al.(2014)Goodfellow, Pouget-Abadie, Mirza, Xu,
  Warde-Farley, Ozair, Courville, and Bengio]{goodfellow2014generative}
Ian Goodfellow, Jean Pouget-Abadie, Mehdi Mirza, Bing Xu, David Warde-Farley,
  Sherjil Ozair, Aaron Courville, and Yoshua Bengio.
\newblock Generative adversarial nets.
\newblock In \emph{Advances in Neural Information Processing Systems},
  volume~27, 2014.
\newblock URL
  \url{https://proceedings.neurips.cc/paper/2014/hash/f033ed80deb0234979a61f95710dbe25-Abstract.html}.

\bibitem[Gu et~al.(2026)Gu, Fang, Jiang, Mao, Han, Cai, and Shou]{anyflow2026}
Yuchao Gu, Guian Fang, Yuxin Jiang, Weijia Mao, Song Han, Han Cai, and
  Mike~Zheng Shou.
\newblock {AnyFlow}: Any-step video diffusion model with on-policy flow map
  distillation.
\newblock \emph{arXiv preprint arXiv:2605.13724}, 2026.
\newblock URL \url{https://arxiv.org/abs/2605.13724}.

\bibitem[Gulrajani et~al.(2017)Gulrajani, Ahmed, Arjovsky, Dumoulin, and
  Courville]{gulrajani2017improved}
Ishaan Gulrajani, Faruk Ahmed, Martin Arjovsky, Vincent Dumoulin, and Aaron~C.
  Courville.
\newblock Improved training of wasserstein {GANs}.
\newblock In \emph{Advances in Neural Information Processing Systems},
  volume~30, 2017.
\newblock URL
  \url{https://proceedings.neurips.cc/paper/2017/hash/892c3b1c6dccd52936e27cbd0ff683d6-Abstract.html}.

\bibitem[Han et~al.(2026)Han, Li, Guo, Xu, Ermon, and Cand\`es]{han2026wflow}
Jiaqi Han, Puheng Li, Qiushan Guo, Renyuan Xu, Stefano Ermon, and Emmanuel~J.
  Cand\`es.
\newblock One-step generative modeling via wasserstein gradient flows.
\newblock \emph{arXiv preprint arXiv:2605.11755}, 2026.
\newblock URL \url{https://arxiv.org/abs/2605.11755}.

\bibitem[He et~al.(2016)He, Zhang, Ren, and Sun]{he2016deep}
Kaiming He, Xiangyu Zhang, Shaoqing Ren, and Jian Sun.
\newblock Deep residual learning for image recognition.
\newblock In \emph{Proceedings of the IEEE Conference on Computer Vision and
  Pattern Recognition}, pp.\  770--778, 2016.

\bibitem[He et~al.(2022)He, Chen, Xie, Li, Doll\'ar, and Girshick]{he2022mae}
Kaiming He, Xinlei Chen, Saining Xie, Yanghao Li, Piotr Doll\'ar, and Ross
  Girshick.
\newblock Masked autoencoders are scalable vision learners.
\newblock In \emph{Proceedings of the IEEE/CVF Conference on Computer Vision
  and Pattern Recognition}, pp.\  16000--16009, 2022.

\bibitem[Heusel et~al.(2017)Heusel, Ramsauer, Unterthiner, Nessler, and
  Hochreiter]{heusel2017gans}
Martin Heusel, Hubert Ramsauer, Thomas Unterthiner, Bernhard Nessler, and Sepp
  Hochreiter.
\newblock {GANs} trained by a two time-scale update rule converge to a local
  {Nash} equilibrium.
\newblock In \emph{Advances in Neural Information Processing Systems},
  volume~30, 2017.
\newblock URL
  \url{https://proceedings.neurips.cc/paper/2017/hash/8a1d694707eb0fefe65871369074926d-Abstract.html}.

\bibitem[Ho \& Salimans(2022)Ho and Salimans]{ho2022classifierfree}
Jonathan Ho and Tim Salimans.
\newblock Classifier-free diffusion guidance.
\newblock \emph{arXiv preprint arXiv:2207.12598}, 2022.
\newblock URL \url{https://arxiv.org/abs/2207.12598}.

\bibitem[Hu et~al.(2022)Hu, Shen, Wallis, Allen-Zhu, Li, Wang, Wang, and
  Chen]{hu2022lora}
Edward~J. Hu, Yelong Shen, Phillip Wallis, Zeyuan Allen-Zhu, Yuanzhi Li, Shean
  Wang, Lu~Wang, and Weizhu Chen.
\newblock {LoRA}: Low-rank adaptation of large language models.
\newblock \emph{International Conference on Learning Representations}, 2022.
\newblock URL \url{https://arxiv.org/abs/2106.09685}.

\bibitem[Huang et~al.(2024)Huang, He, Yu, Zhang, Si, Jiang, Zhang, Wu, Jin,
  Chanpaisit, Wang, Chen, Wang, Lin, Qiao, and Liu]{huang2024vbench}
Ziqi Huang, Yinan He, Jiashuo Yu, Fan Zhang, Chenyang Si, Yuming Jiang, Yuanhan
  Zhang, Tianxing Wu, Qingyang Jin, Nattapol Chanpaisit, Yaohui Wang, Xinyuan
  Chen, Limin Wang, Dahua Lin, Yu~Qiao, and Ziwei Liu.
\newblock {VBench}: Comprehensive benchmark suite for video generative models.
\newblock \emph{arXiv preprint arXiv:2311.17982}, 2024.
\newblock URL \url{https://arxiv.org/abs/2311.17982}.

\bibitem[Kang et~al.(2023)Kang, Zhu, Zhang, Park, Shechtman, Paris, and
  Park]{kang2023gigagan}
Minguk Kang, Jun-Yan Zhu, Richard Zhang, Jaesik Park, Eli Shechtman, Sylvain
  Paris, and Taesung Park.
\newblock Scaling up {GANs} for text-to-image synthesis.
\newblock \emph{arXiv preprint arXiv:2303.05511}, 2023.
\newblock URL \url{https://arxiv.org/abs/2303.05511}.

\bibitem[Kim et~al.(2024)Kim, Lai, Liao, Takida, Murata, Uesaka, Mitsufuji, and
  Ermon]{kim2024pagoda}
Dongjun Kim, Chieh-Hsin Lai, Wei-Hsiang Liao, Yuhta Takida, Naoki Murata,
  Toshimitsu Uesaka, Yuki Mitsufuji, and Stefano Ermon.
\newblock {PaGoDA}: Progressive growing of a one-step generator from a
  low-resolution diffusion teacher.
\newblock In \emph{Advances in Neural Information Processing Systems}, 2024.
\newblock URL \url{https://arxiv.org/abs/2405.14822}.

\bibitem[Kingma \& Gao(2023)Kingma and Gao]{kingma2023vdmpp}
Diederik~P. Kingma and Ruiqi Gao.
\newblock Understanding diffusion objectives as the {ELBO} with simple data
  augmentation.
\newblock In \emph{Advances in Neural Information Processing Systems}, 2023.
\newblock URL \url{https://arxiv.org/abs/2303.00848}.

\bibitem[Li \& He(2025)Li and He]{li2025jit}
Tianhong Li and Kaiming He.
\newblock Back to basics: Let denoising generative models denoise.
\newblock \emph{arXiv preprint arXiv:2511.13720}, 2025.
\newblock URL \url{https://arxiv.org/abs/2511.13720}.

\bibitem[Li et~al.(2024)Li, Tian, Li, Deng, and He]{li2024mar}
Tianhong Li, Yonglong Tian, He~Li, Mingyang Deng, and Kaiming He.
\newblock Autoregressive image generation without vector quantization.
\newblock In \emph{Advances in Neural Information Processing Systems}, 2024.
\newblock URL \url{https://arxiv.org/abs/2406.11838}.

\bibitem[Lipman et~al.(2023)Lipman, Chen, Ben-Hamu, Nickel, and
  Le]{lipman2023flow}
Yaron Lipman, Ricky T.~Q. Chen, Heli Ben-Hamu, Maximilian Nickel, and Matt Le.
\newblock Flow matching for generative modeling.
\newblock In \emph{International Conference on Learning Representations}, 2023.
\newblock URL \url{https://openreview.net/forum?id=PqvMRDCJT9t}.

\bibitem[Loshchilov \& Hutter(2019)Loshchilov and
  Hutter]{loshchilov2019decoupled}
Ilya Loshchilov and Frank Hutter.
\newblock Decoupled weight decay regularization.
\newblock In \emph{International Conference on Learning Representations}, 2019.
\newblock URL \url{https://openreview.net/forum?id=Bkg6RiCqY7}.

\bibitem[Ma et~al.(2024)Ma, Goldstein, Albergo, Boffi, Vanden-Eijnden, and
  Xie]{ma2024sit}
Nanye Ma, Mark Goldstein, Michael~S. Albergo, Nicholas~M. Boffi, Eric
  Vanden-Eijnden, and Saining Xie.
\newblock {SiT}: Exploring flow and diffusion-based generative models with
  scalable interpolant transformers.
\newblock In \emph{European Conference on Computer Vision}, 2024.
\newblock URL \url{https://arxiv.org/abs/2401.08740}.

\bibitem[Miyato et~al.(2018)Miyato, Kataoka, Koyama, and
  Yoshida]{miyato2018spectral}
Takeru Miyato, Toshiki Kataoka, Masanori Koyama, and Yuichi Yoshida.
\newblock Spectral normalization for generative adversarial networks.
\newblock In \emph{International Conference on Learning Representations}, 2018.
\newblock URL \url{https://openreview.net/forum?id=B1QRgziT-}.

\bibitem[Mroueh et~al.(2018)Mroueh, Li, Sercu, Raj, and
  Cheng]{mroueh2018sobolev}
Youssef Mroueh, Chun-Liang Li, Tom Sercu, Anant Raj, and Yu~Cheng.
\newblock Sobolev {GAN}.
\newblock In \emph{International Conference on Learning Representations}, 2018.
\newblock URL \url{https://openreview.net/forum?id=r1f78iAcY7}.

\bibitem[Peebles \& Xie(2023)Peebles and Xie]{peebles2023dit}
William Peebles and Saining Xie.
\newblock Scalable diffusion models with transformers.
\newblock In \emph{Proceedings of the IEEE/CVF International Conference on
  Computer Vision}, 2023.
\newblock URL \url{https://arxiv.org/abs/2212.09748}.

\bibitem[Salimans et~al.(2016)Salimans, Goodfellow, Zaremba, Cheung, Radford,
  and Chen]{salimans2016improved}
Tim Salimans, Ian Goodfellow, Wojciech Zaremba, Vicki Cheung, Alec Radford, and
  Xi~Chen.
\newblock Improved techniques for training {GANs}.
\newblock In \emph{Advances in Neural Information Processing Systems},
  volume~29, 2016.
\newblock URL
  \url{https://proceedings.neurips.cc/paper/2016/hash/8a3363abe792db2d8761d6403605aeb7-Abstract.html}.

\bibitem[Sauer et~al.(2021)Sauer, Chitta, M{\"u}ller, and
  Geiger]{sauer2021projected}
Axel Sauer, Kashyap Chitta, Jens M{\"u}ller, and Andreas Geiger.
\newblock Projected {GANs} converge faster.
\newblock In \emph{Advances in Neural Information Processing Systems},
  volume~34, pp.\  17480--17492. Curran Associates, Inc., 2021.
\newblock URL
  \url{https://proceedings.neurips.cc/paper/2021/hash/9219adc5c42107c4911e249155320648-Abstract.html}.

\bibitem[Sauer et~al.(2022)Sauer, Schwarz, and Geiger]{sauer2022styleganxl}
Axel Sauer, Katja Schwarz, and Andreas Geiger.
\newblock {StyleGAN-XL}: Scaling {StyleGAN} to large diverse datasets.
\newblock In \emph{ACM SIGGRAPH 2022 Conference Proceedings}, 2022.
\newblock \doi{10.1145/3528233.3530738}.
\newblock URL \url{https://arxiv.org/abs/2202.00273}.

\bibitem[Sim\'eoni et~al.(2025)Sim\'eoni, Vo, Seitzer, Baldassarre, Oquab,
  Jose, Khalidov, Szafraniec, Yi, Ramamonjisoa, Massa, Haziza, Wehrstedt, Wang,
  Darcet, Moutakanni, Sentana, Roberts, Vedaldi, Tolan, Brandt, Couprie,
  Mairal, J\'egou, Labatut, and Bojanowski]{simeoni2025dinov3}
Oriane Sim\'eoni, Huy~V. Vo, Maximilian Seitzer, Federico Baldassarre, Maxime
  Oquab, Cijo Jose, Vasil Khalidov, Marc Szafraniec, Seungeun Yi, Micha\"el
  Ramamonjisoa, Francisco Massa, Daniel Haziza, Luca Wehrstedt, Jianyuan Wang,
  Timoth\'ee Darcet, Th\'eo Moutakanni, Leonel Sentana, Claire Roberts, Andrea
  Vedaldi, Jamie Tolan, John Brandt, Camille Couprie, Julien Mairal, Herv\'e
  J\'egou, Patrick Labatut, and Piotr Bojanowski.
\newblock {DINOv3}.
\newblock \emph{arXiv preprint arXiv:2508.10104}, 2025.
\newblock URL \url{https://arxiv.org/abs/2508.10104}.

\bibitem[Song et~al.(2021)Song, Sohl-Dickstein, Kingma, Kumar, Ermon, and
  Poole]{song2021score}
Yang Song, Jascha Sohl-Dickstein, Diederik~P. Kingma, Abhishek Kumar, Stefano
  Ermon, and Ben Poole.
\newblock Score-based generative modeling through stochastic differential
  equations.
\newblock In \emph{International Conference on Learning Representations}, 2021.
\newblock URL \url{https://openreview.net/forum?id=PxTIG12RRHS}.

\bibitem[Sun et~al.(2025)Sun, Jiang, and Lin]{sun2025unified}
Peng Sun, Yi~Jiang, and Tao Lin.
\newblock Unified continuous generative models.
\newblock \emph{arXiv preprint arXiv:2505.07447}, 2025.
\newblock URL \url{https://arxiv.org/abs/2505.07447}.

\bibitem[Sun et~al.(2026)Sun, Cheng, Liu, Xie, Shang, and
  Lin]{sun2026threebody}
Peng Sun, Zhenglin Cheng, Deyuan Liu, Jun Xie, Xinyi Shang, and Tao Lin.
\newblock Three-body scattering for generative modeling.
\newblock \emph{arXiv preprint arXiv:2607.18198}, 2026.
\newblock URL \url{https://arxiv.org/abs/2607.18198}.

\bibitem[Tian et~al.(2024)Tian, Jiang, Yuan, Peng, and Wang]{tian2024var}
Keyu Tian, Yi~Jiang, Zehuan Yuan, Bingyue Peng, and Liwei Wang.
\newblock Visual autoregressive modeling: Scalable image generation via
  next-scale prediction.
\newblock \emph{arXiv preprint arXiv:2404.02905}, 2024.
\newblock URL \url{https://arxiv.org/abs/2404.02905}.

\bibitem[Tong et~al.(2022)Tong, Song, Wang, and Wang]{tong2022videomae}
Zhan Tong, Yibing Song, Jue Wang, and Limin Wang.
\newblock {VideoMAE}: Masked autoencoders are data-efficient learners for
  self-supervised video pre-training.
\newblock In \emph{Advances in Neural Information Processing Systems}, 2022.

\bibitem[Tschannen et~al.(2025)Tschannen, Gritsenko, Wang, Naeem,
  Alabdulmohsin, Parthasarathy, Evans, Beyer, Xia, Mustafa, H\'enaff, Harmsen,
  Steiner, and Zhai]{tschn2025siglip2}
Michael Tschannen, Alexey Gritsenko, Xiao Wang, Muhammad~Ferjad Naeem, Ibrahim
  Alabdulmohsin, Nikhil Parthasarathy, Talfan Evans, Lucas Beyer, Ye~Xia, Basil
  Mustafa, Olivier H\'enaff, Jeremiah Harmsen, Andreas Steiner, and Xiaohua
  Zhai.
\newblock {SigLIP 2}: Multilingual vision-language encoders with improved
  semantic understanding, localization, and dense features.
\newblock \emph{arXiv preprint arXiv:2502.14786}, 2025.

\bibitem[Villani(2009)]{villani2009optimal}
C{\'e}dric Villani.
\newblock \emph{Optimal Transport: Old and New}, volume 338 of
  \emph{Grundlehren der mathematischen Wissenschaften}.
\newblock Springer, 2009.
\newblock \doi{10.1007/978-3-540-71050-9}.

\bibitem[{Wan Team}(2025)]{wan2025wan}
{Wan Team}.
\newblock Wan: Open and advanced large-scale video generative models.
\newblock \emph{arXiv preprint arXiv:2503.20314}, 2025.
\newblock URL \url{https://arxiv.org/abs/2503.20314}.

\bibitem[Wang et~al.(2025)Wang, Zhu, Xu, Yang, Lin, Yan, Wang, Wang, and
  Wang]{wang2025internvideonext}
Chenting Wang, Yuhan Zhu, Yicheng Xu, Jiange Yang, Lang Lin, Ziang Yan, Yali
  Wang, Yi~Wang, and Limin Wang.
\newblock {InternVideo-Next}: Towards general video foundation models without
  video-text supervision.
\newblock \emph{arXiv preprint arXiv:2512.01342}, 2025.

\bibitem[Yang et~al.(2026)Yang, Geng, Ju, Tian, and Wang]{yang2026fdloss}
Jiawei Yang, Zhengyang Geng, Xuan Ju, Yonglong Tian, and Yue Wang.
\newblock Representation {Fr\'{e}chet} loss for visual generation.
\newblock \emph{arXiv preprint arXiv:2604.28190}, 2026.
\newblock URL \url{https://arxiv.org/abs/2604.28190}.

\bibitem[Yao \& Wang(2025)Yao and Wang]{yao2025reconstruction}
Jingfeng Yao and Xinggang Wang.
\newblock Reconstruction vs. generation: Taming optimization dilemma in latent
  diffusion models.
\newblock \emph{arXiv preprint arXiv:2501.01423}, 2025.
\newblock URL \url{https://arxiv.org/abs/2501.01423}.

\bibitem[Yin et~al.(2024)Yin, Gharbi, Zhang, Shechtman, Durand, Freeman, and
  Park]{yin2024improved}
Tianwei Yin, Micha{\"e}l Gharbi, Richard Zhang, Eli Shechtman, Fr{\'e}do
  Durand, William~T. Freeman, and Taesung Park.
\newblock Improved distribution matching distillation for fast image synthesis.
\newblock \emph{arXiv preprint arXiv:2405.14867}, 2024.
\newblock URL \url{https://arxiv.org/abs/2405.14867}.

\bibitem[Yu et~al.(2026)Yu, Liu, He, Zhang, Liu, Chen, and Chen]{yu2026bar}
Qihang Yu, Qihao Liu, Ju~He, Xinyang Zhang, Yang Liu, Liang-Chieh Chen, and
  Xi~Chen.
\newblock Autoregressive image generation with masked bit modeling.
\newblock \emph{arXiv preprint arXiv:2602.09024}, 2026.
\newblock URL \url{https://arxiv.org/abs/2602.09024}.

\bibitem[Zheng et~al.(2025)Zheng, Wang, Ma, Chen, Zhang, Balaji, Chen, Liu,
  Zhu, and Zhang]{rcm2025}
Kaiwen Zheng, Yuji Wang, Qianli Ma, Huayu Chen, Jintao Zhang, Yogesh Balaji,
  Jianfei Chen, Ming-Yu Liu, Jun Zhu, and Qinsheng Zhang.
\newblock Large scale diffusion distillation via score-regularized
  continuous-time consistency.
\newblock \emph{arXiv preprint arXiv:2510.08431}, 2025.
\newblock URL \url{https://arxiv.org/abs/2510.08431}.

\bibitem[Zoran et~al.(2025)Zoran, Parthasarathy, Yang, Hudson, Carreira, and
  Zisserman]{zoran2025rvm}
Daniel Zoran, Nikhil Parthasarathy, Yi~Yang, Drew~A. Hudson, Joao Carreira, and
  Andrew Zisserman.
\newblock Recurrent video masked autoencoders.
\newblock \emph{arXiv preprint arXiv:2512.13684}, 2025.

\end{thebibliography}
\bibliographystyle{configuration/iclr2027_conference}
\appendix

\section{Implementation details}
\label{app:implementation}
\paragraph{Image experiments.}
ImageNet experiments use frozen MAE-B/16, SigLIP2-B/16, and ResNet-18 encoders~\citep{he2022mae,tschn2025siglip2,he2016deep} with independent condition-aware potential heads. JiT-B is initialized from the released JiT checkpoint~\citep{li2025jit}; DiT-XL/1~\citep{peebles2023dit} and the VA-VAE decoder~\citep{yao2025reconstruction} follow the UCGM release~\citep{sun2025unified}. The main DiT field-regression run uses 100K updates on eight H800 GPUs, while the controlled JiT-B ablations use 20K updates. We optimize the generator and potential heads with AdamW~\citep{loshchilov2019decoupled} ($\beta_1=0.9,\beta_2=0.999$) at learning rates of $10^{-5}$ and $5\times10^{-5}$, respectively. We use two critic updates per generator update, a per-GPU batch size of 32 (global batch size 256 for the eight-GPU run), and a generator EMA decay of 0.999. MAE and SigLIP2 inputs are bicubically resized to 224 and 256 pixels, respectively; ResNet-18 inputs are bilinearly resized to 224 pixels. Each encoder then applies its checkpoint-specific channel normalization. Evaluation uses 50K generated images for both the main DiT run and the controlled JiT-B ablations. The 100K field-regression run consumed 158.3 H800 GPU-hours, including two checkpoint restarts (19.78 hours of eight-GPU Slurm allocation); its final segment averaged 0.669 seconds per update. Table~\ref{tab:imagenet} reports nominal JiT-B/16 (131M) and standard DiT-XL/2 (675M) backbone sizes~\citep{li2025jit,peebles2023dit}. Our instantiated JiT-B and DiT-XL/1 generators have approximately 131.3M and 676.1M trainable parameters, respectively, excluding critics and representers.

\paragraph{Displacement normalization.}
For each representer channel, we center and standardize tokens using exponential-moving-average first and second moments. We then normalize each token to Euclidean norm $\sqrt d$, where $d$ is its channel dimension. The field-regression run minimizes $\frac{\lambda}{2}\E\|\nabla_{\xx}u_{\mphi}(\xx_\alpha,\cc)-\mDelta/\lambda\|_2^2$ with $\lambda=5$; the norm is summed over token and channel dimensions before batch averaging. Its supervised target is therefore $\mDelta/5$. Expanding the square gives the critic loss $-\E[u_{\mphi}(\xx_{\mathrm{r}},\cc)-u_{\mphi}(\xx_{\mathrm{p}},\cc)]+\frac{\lambda}{2}\E\|\nabla_{\xx}u_{\mphi}(\xx_\alpha,\cc)\|_2^2$ plus a term independent of $\mphi$. For an unrestricted potential, setting $\widetilde u=\lambda u$ recovers the unit-weight regression up to a positive factor; this does not assert identical optimization under the spectral normalization used for the potential heads.

\paragraph{Video experiments.}
Both model sizes use the Wan2.1-14B 480p 250K subset of the Wan\_datasets released with rCM~\citep{rcm2025}. This dataset contains videos generated offline by Wan2.1-14B; these serve as the reference (``real'') samples for video training. Optimization requires no online teacher scores or trajectories.

Wan2.1 is fine-tuned with LoRA~\citep{hu2022lora} of rank 256 and scaling parameter $\alpha_{\mathrm{LoRA}}=256$, applied to the attention projections and feed-forward layers. With the adapters disabled, we extract text-conditioned features from clean video latents at blocks 7, 11, and 15 (zero-based indices), using a fixed model timestep of 500 without adding noise to the latents. The features undergo $2\times2$ spatial average pooling, preserving the temporal dimension, and each selected layer feeds an independent text-conditioned potential head. The base-model weights remain frozen, while gradients propagate through its features to the generator. Backward simulation uses the Wan velocity parameterization.

\paragraph{Evaluation.}
Wan2.1-14B is evaluated with $\sigma_{\max}=1600$, CFG$=1$, and seed 0 at one, two, and four steps. Each budget uses 944 augmented VBench prompts with five samples per prompt (4,720 videos), at $832\times480\times81$ and 16 fps. All 16 VBench dimensions are evaluated.

We report FID~\citep{heusel2017gans} and Inception Score (IS)~\citep{salimans2016improved} for ImageNet generation; sample counts for the main DiT run and controlled JiT-B ablations are specified above.

\subsection{Algorithm and code for the equivalent adversarial form}
\label{app:gan-code}
We use the unit-weight convention of \algxref{alg:nextgan} throughout this subsection. For a fixed generator, expanding the regression loss and applying Eq.~\eqref{eq:line-integral} gives
\begin{equation}
\frac{1}{2}\E\|\nabla_{\xx}u_{\mphi}(\xx_\alpha,\cc)-\mDelta\|_2^2
=\underbrace{\E[u_{\mphi}(\xx_{\mathrm{p}},\cc)-u_{\mphi}(\xx_{\mathrm{r}},\cc)]
+\frac{1}{2}\E\|\nabla_{\xx}u_{\mphi}(\xx_\alpha,\cc)\|_2^2}_{\cL_{\mathrm{D}}^{\mathrm{adv}}}
+\frac{1}{2}\E\|\mDelta\|_2^2.
\end{equation}
The last term is independent of the critic parameters, and $\cL_{\mathrm{D}}^{\mathrm{adv}}$ is the negative of the value in Eq.~\eqref{eq:nextgan-value}. The equivalence holds after averaging over uniform interpolation times; the two finite-minibatch estimators need not coincide. The penalty is squared gradient energy, not the unit-norm penalty $(\|\nabla u\|_2-1)^2$ used in WGAN-GP.

\algxref{alg:nextgan-gan} writes the alternating updates in adversarial form for one frozen representer and one scalar critic. Each displayed loss is averaged over the minibatch; real and generated samples are paired independently conditional on $\cc$. The critic update detaches the endpoints from the generator, but differentiates the gradient-energy term with respect to $\mphi$. The generator update holds $\mphi$ fixed while retaining input gradients through both the critic and the frozen representer. For multiple representers, sum the corresponding losses.

\begin{algorithm}[H]
\caption{\NextGAN training in adversarial form}
\label{alg:nextgan-gan}
\begin{algorithmic}[1]
\Require frozen representer $\mf$; critic steps per generator step $n_{\mathrm{critic}}$
\State Initialize generator $\mg_{\mtheta}$ and potential $u_{\mphi}$
\Repeat
    \State Sample $(\bar{\xx}_{\mathrm{r}},\cc)$, $\zz\sim p(\zz)$, and generate $\bar{\xx}_{\mathrm{p}}\gets\mg_{\mtheta}(\zz,\cc)$
    \State Compute $\xx_{\mathrm{r}}\gets\mf(\bar{\xx}_{\mathrm{r}},\cc)$ and $\xx_{\mathrm{p}}\gets\mf(\bar{\xx}_{\mathrm{p}},\cc)$
    \For{$k=1,\ldots,n_{\mathrm{critic}}$}
        \State Sample $\alpha\sim\mathcal{U}(0,1)$ and set $\xx_\alpha\gets\alpha\xx_{\mathrm{r}}+(1-\alpha)\sg[\xx_{\mathrm{p}}]$
        \State Update $\mphi$ by descending $\displaystyle
            u_{\mphi}(\sg[\xx_{\mathrm{p}}],\cc)-u_{\mphi}(\xx_{\mathrm{r}},\cc)
            +\frac{1}{2}\left\|\nabla_{\xx}u_{\mphi}(\xx_\alpha,\cc)\right\|_2^2$
    \EndFor
    \State Hold $\mphi$ fixed; update $\mtheta$ by descending $\displaystyle -u_{\mphi}(\xx_{\mathrm{p}},\cc)$
\Until{convergence}
\end{algorithmic}
\end{algorithm}

The following PyTorch code implements one alternating update for one frozen representer and one scalar critic, using the endpoint-detachment convention of \algxref{alg:nextgan}. The generator loss remains $-\E[u_{\mphi}(\mf(\mg_{\mtheta}(\zz,\cc),\cc),\cc)]$. The representer includes feature preprocessing and normalization; its parameters are frozen, but input gradients are retained during the generator update. For multiple representers, sum their critic losses before the critic update and their generator losses before the generator update. Generator EMA and distributed-training plumbing are omitted.
\begin{lstlisting}[language=Python,basicstyle=\ttfamily\scriptsize,
    keywordstyle=\bfseries,commentstyle=\color{gray},
    columns=fullflexible,keepspaces=true,showstringspaces=false,
    frame=single,rulecolor=\color{lightgray},breaklines=true,
    firstline=3]
import torch


def nextgan_step(G, F, D, opt_g, opt_d, real, z, c,
                 n_critic=2):
    # F includes feature normalization; D returns one scalar per sample.
    # F and D must not couple different samples in a minibatch.
    F.requires_grad_(False)
    F.eval()
    D.requires_grad_(True)
    with torch.no_grad():
        xr, xp = F(real, c), F(G(z, c), c)

    for _ in range(n_critic):
        shape = (xr.shape[0],) + (1,) * (xr.ndim - 1)
        alpha = torch.rand(shape, device=xr.device, dtype=xr.dtype)
        xa = ((1 - alpha) * xp + alpha * xr).requires_grad_(True)
        grad = torch.autograd.grad(D(xa, c).sum(), xa,
                                   create_graph=True)[0]
        energy = grad.flatten(1).square().sum(1).mean()
        loss_d = D(xp, c).mean() - D(xr, c).mean() + 0.5 * energy
        opt_d.zero_grad(set_to_none=True)
        loss_d.backward()
        opt_d.step()

    D.requires_grad_(False)
    opt_g.zero_grad(set_to_none=True)
    # Keep input gradients through the frozen representer and critic.
    loss_g = -D(F(G(z, c), c), c).mean()
    loss_g.backward()
    opt_g.step()
    D.requires_grad_(True)
    return loss_d.detach(), loss_g.detach()
\end{lstlisting}

\section{Ablations: training objective and feature space}
\label{sec:mechanism}
\label{app:ablations}

We examine training objectives and feature spaces using JiT-B. Both the training curves in \figref{fig:mechanism-signal}a and the representer ablation in panel b report FID-50K using the JiT evaluator.

\begin{figure}[H]
\centering
\begingroup
\newsavebox{\ablationplotbox}
\newsavebox{\ablationtablebox}
\newlength{\ablationpanelheight}
\newlength{\ablationpaneltopoffset}
\definecolor{ablationnextgan}{HTML}{D97864}
\definecolor{ablationlogistic}{HTML}{5597B0}
\definecolor{ablationwgan}{HTML}{8971AA}
\definecolor{ablationtbsm}{HTML}{C8993E}
\pgfplotstableread[col sep=comma]{figures/jit_complete_curves/jit_curves_displayed_50k.csv}\ablationfiddata
\newcommand{\ablationfidcurve}[3]{%
  \addplot+[#2,
    x filter/.code={%
      \pgfplotstablegetelem{\coordindex}{method}\of\ablationfiddata
      \edef\currentrun{\pgfplotsretval}%
      \def\selectedrun{#1}%
      \ifx\currentrun\selectedrun\else\def\pgfmathresult{nan}\fi
    }]
    table[x expr=\thisrow{step}/1000,y=fid_jit] {\ablationfiddata};
  \addlegendentry{#3}
}
\noindent\makebox[0.50\linewidth][l]{\textbf{a}}\hfill
\makebox[0.42\linewidth][l]{\textbf{b}}\par\smallskip
\noindent\begin{minipage}[t]{0.50\linewidth}
\vspace{0pt}
\centering
\begin{lrbox}{\ablationplotbox}
\begin{tikzpicture}
  \begin{axis}[
    name=ablationaxis,
    width=\linewidth, height=5.8cm,
    xmin=-1, xmax=56, ymin=2.6, ymax=380, ymode=log,
    xtick={0,10,20,30,40,50},
    ytick={3,5,10,20,50,100,300},
    yticklabels={3,5,10,20,50,100,300},
    minor y tick num=0,
    xlabel={Generator updates (K)}, ylabel={FID-50K},
    tick label style={font=\scriptsize},
    label style={font=\footnotesize},
    axis lines=left, axis line style={gray!65},
    tick style={gray!65}, tick align=outside,
    ymajorgrids=true, grid style={gray!16,line width=0.3pt},
    every axis plot/.append style={line width=0.8pt,mark size=1.3pt},
    unbounded coords=discard, filter discard warning=false,
    legend style={at={(0.98,0.98)},anchor=north east,
      draw=none,fill=white,font=\scriptsize,
      cells={anchor=west},row sep=0.5pt,inner sep=2pt},
    legend image post style={line width=0.8pt},
    clip=false,
  ]
    \ablationfidcurve{NextGAN (adversarial form)}{color=ablationnextgan,mark=*,mark options={fill=ablationnextgan,draw=white,line width=0.3pt}}{\textsc{NextGAN} (adversarial form)}
    \ablationfidcurve{GAN}{color=ablationlogistic,mark=square*,mark options={fill=ablationlogistic,draw=white,line width=0.3pt}}{GAN}
    \ablationfidcurve{WGAN (no GP)}{color=ablationwgan,mark=triangle*,mark options={fill=ablationwgan,draw=white,line width=0.3pt}}{WGAN (no GP)}
    \ablationfidcurve{TBSM GAN corner}{color=ablationtbsm,mark=diamond*,mark options={fill=ablationtbsm,draw=white,line width=0.3pt}}{TBSM GAN corner}
    \node[anchor=west,inner sep=1pt,font=\tiny,text=ablationnextgan] at (axis cs:50,3.314906) {3.31};
    \node[anchor=west,inner sep=1pt,font=\tiny,text=ablationlogistic] at (axis cs:50,5.644214) {5.64};
    \node[anchor=west,inner sep=1pt,font=\tiny,text=ablationwgan] at (axis cs:50,17.129610) {17.13};
    \node[anchor=west,inner sep=1pt,font=\tiny,text=ablationtbsm] at (axis cs:50,28.694857) {28.69};
  \end{axis}
  \pgfextracty{\ablationpanelheight}{\pgfpointdiff
    {\pgfpointanchor{ablationaxis}{south west}}
    {\pgfpointanchor{ablationaxis}{north west}}}
  \global\ablationpanelheight=\ablationpanelheight
  \pgfextracty{\ablationpaneltopoffset}{\pgfpointdiff
    {\pgfpointanchor{ablationaxis}{north west}}
    {\pgfpointanchor{current bounding box}{north}}}
  \global\ablationpaneltopoffset=\ablationpaneltopoffset
\end{tikzpicture}
\end{lrbox}%
\noindent\makebox[\linewidth][l]{\usebox{\ablationplotbox}}
\end{minipage}\hfill
\begin{minipage}[t]{0.42\linewidth}
\vspace*{\ablationpaneltopoffset}
\footnotesize
\setlength{\tabcolsep}{2pt}
\renewcommand{\arraystretch}{1}
\newcommand{\ablationrepresentertable}{%
\begin{tabularx}{\linewidth}{@{}l*{3}{>{\centering\arraybackslash}X}r@{}}
\toprule
 & \scriptsize MAE & \scriptsize SigLIP2 & \scriptsize ResNet-18 & FID-50K $\downarrow$ \\
\midrule
\textsc{Single} & $\checkmark$ & -- & -- & 7.08 \\
 & -- & $\checkmark$ & -- & 17.78 \\
 & -- & -- & $\checkmark$ & 5.55 \\
\midrule
\textsc{Pair} & $\checkmark$ & $\checkmark$ & -- & 6.64 \\
 & $\checkmark$ & -- & $\checkmark$ & 4.67 \\
 & -- & $\checkmark$ & $\checkmark$ & 4.65 \\
\midrule
\rowcolor{nextganpale}
\textsc{All} & $\checkmark$ & $\checkmark$ & $\checkmark$ & \textbf{4.10} \\
\bottomrule
\end{tabularx}%
}
\setlength{\extrarowheight}{0pt}
\sbox{\ablationtablebox}{\ablationrepresentertable}
\setlength{\extrarowheight}{\dimexpr\ablationpanelheight-\ht\ablationtablebox-\dp\ablationtablebox\relax}
\divide\extrarowheight by 8
\sbox{\ablationtablebox}{\ablationrepresentertable}
\typeout{Ablation panel heights: data rectangle=\the\ablationpanelheight; table=\the\dimexpr\ht\ablationtablebox+\dp\ablationtablebox\relax}
\noindent\makebox[\linewidth][r]{\usebox{\ablationtablebox}}
\end{minipage}\par
\endgroup
\small\caption{\textbf{Training objective and feature space.} \emph{(a)} JiT-B training curves through 50K updates (logarithmic FID-50K axis). \emph{(b)} Frozen representer ablations on JiT-B at 20K updates. Lower is better.}
\label{fig:mechanism-signal}
\end{figure}

\paragraph{Training objective.}
\figref{fig:mechanism-signal}a shows \NextGAN (adversarial form), logistic GAN, WGAN without gradient penalty, and the TBSM GAN corner in the JiT-B representation-space setting. At 50K updates, their FID-50K values are 3.31, 5.64, 17.13, and 28.69, respectively. The plotted TBSM curve combines a same-configuration independent run at 0--2.5K with the original run at 4--50K; at the overlapping 2K checkpoint, their FID-50K values are 16.26 and 16.01, respectively. Each curve includes the measured shared-initialization FID at step zero using 50K generated images. Lines connecting checkpoints are visual guides.

\paragraph{Feature space.}
\figref{fig:mechanism-signal}b compares frozen representers after 20K updates. Individual representers yield FID-50K values from 5.55 to 17.78. Combining all three reaches 4.10, a 26\% reduction from the best single representer, and also improves over every tested pair. These results support using complementary feature spaces to provide the critic with richer training signals.

\FloatBarrier
\section{Proof of \texorpdfstring{\propref{prop:population-field}}{Prop. \getrefnumber{prop:population-field}}}
\label{app:population-field}
Let $\mY=(\mX_\alpha,\mC)$, let $\mDelta=\mX_{\mathrm{r}}-\mX_{\mathrm{p}}$, and write $\langle\vv_1,\vv_2\rangle=\E\langle\vv_1(\mY),\vv_2(\mY)\rangle$ for the inner product of $L^2(\mu_{\mtheta})$.

\paragraph{Part 1.}
By definition, $\vv_{\mtheta}(\mY)=\E[\mDelta\mid\mY]$. For any square-integrable field $\ww$,
\begin{equation}
    \frac12\E\|\ww(\mY)-\mDelta\|_2^2
    =\frac12\|\ww-\vv_{\mtheta}\|^2
    +\frac12\E\|\mDelta-\vv_{\mtheta}(\mY)\|_2^2,
    \label{eq:regression-decomposition}
\end{equation}
since the cross term $\E\langle \ww(\mY)-\vv_{\mtheta}(\mY),\,\E[\vv_{\mtheta}(\mY)-\mDelta\mid\mY]\rangle$ vanishes. The second term does not depend on $\ww$, so the unrestricted minimizer is $\vv_{\mtheta}$, and the minimizer over the closed subspace $\cG$ is $\operatorname{Proj}_{\cG}\vv_{\mtheta}$.

\paragraph{Part 2.}
For $u\in\cH$, Eq.~\eqref{eq:line-integral}, Fubini's theorem, and the tower property give
\begin{equation}
    \E\big[u(\mX_{\mathrm{r}},\mC)-u(\mX_{\mathrm{p}},\mC)\big]
    =\E\langle\nabla_{\xx}u(\mX_\alpha,\mC),\mDelta\rangle
    =\langle\nabla_{\xx}u,\vv_{\mtheta}\rangle.
    \label{eq:population-line-integral}
\end{equation}
Hence the objective in Eq.~\eqref{eq:distribution-objective} equals $\langle\ww,\vv_{\mtheta}\rangle-\frac12\|\ww\|^2$ with $\ww=\nabla_{\xx}u$. This functional is continuous in $\ww$, so its supremum over $\{\nabla_{\xx}u:u\in\cH\}$ equals its supremum over the closure $\cG$, which is a closed linear subspace because $\cH$ is linear. Writing $\ww^\star=\operatorname{Proj}_{\cG}\vv_{\mtheta}$, we have $\langle\ww,\vv_{\mtheta}\rangle=\langle\ww,\ww^\star\rangle$ for $\ww\in\cG$, and
\begin{equation}
    \langle\ww,\ww^\star\rangle-\frac12\|\ww\|^2
    =\frac12\|\ww^\star\|^2-\frac12\|\ww-\ww^\star\|^2,
\end{equation}
which is maximized at $\ww=\ww^\star$. Therefore $\Dnext=\frac12\|\operatorname{Proj}_{\cG}\vv_{\mtheta}\|^2\ge0$.

\paragraph{Part 3.}
If $P_{\mtheta}(\cdot\mid\cc)=Q(\cdot\mid\cc)$ for almost every $\cc$, the joint law of $(\mX_{\mathrm{r}},\mX_{\mathrm{p}},\alpha,\mC)$ is invariant under $(\mX_{\mathrm{r}},\mX_{\mathrm{p}},\alpha)\mapsto(\mX_{\mathrm{p}},\mX_{\mathrm{r}},1-\alpha)$. This map leaves $\mX_\alpha$ unchanged and negates $\mDelta$. Hence the conditional law of $\mDelta$ given $(\mX_\alpha,\mC)$ is symmetric about zero, so $\vv_{\mtheta}=\zero$ almost surely and $\Dnext=0$ by Part 2.

Conversely, suppose $\Dnext=0$ and fix $u\in\cH$. Since $\cH$ is linear, $tu\in\cH$ for every $t\in\R$, and
\begin{equation}
    0\ \ge\ t\,\langle\nabla_{\xx}u,\vv_{\mtheta}\rangle-\frac{t^2}{2}\|\nabla_{\xx}u\|^2
    \qquad\text{for all } t\in\R.
\end{equation}
Letting $t\to0^{\pm}$ gives $\langle\nabla_{\xx}u,\vv_{\mtheta}\rangle=0$, so by Eq.~\eqref{eq:population-line-integral}, $\E[u(\mX_{\mathrm{r}},\mC)]=\E[u(\mX_{\mathrm{p}},\mC)]$ for every $u\in\cH$. Because $\cH$ determines the joint law, these equalities imply equality of the real and generated joint laws of $(\mX,\mC)$. The product test functions in the proposition are one sufficient choice of such a class. Since both share the marginal of $\mC$, their conditional laws agree for almost every $\cc$. The argument uses independent real--generated sampling and requires no optimal-transport coupling.

\section{Negative-Sobolev form (\texorpdfstring{\remarkref{rem:sobolev}}{Rem. \getrefnumber{rem:sobolev}})}
\label{app:sobolev}
Fix $\cc$ and write $\mu=\mu_{\mtheta}(\cdot\mid\cc)$. Let $q$ and $p_{\mtheta}$ be the densities of $Q(\cdot\mid\cc)$ and $P_{\mtheta}(\cdot\mid\cc)$, respectively, and let $\omega$ be the density of $\mu$. For unrestricted $u$, the objective $J(u)=\int u(q-p_{\mtheta})\,\dm\xx-\frac12\int\|\nabla_{\xx}u\|^2\dm\mu$ is concave. Its first variation in direction $\varphi$ is $\int\varphi(q-p_{\mtheta})\,\dm\xx-\int\langle\nabla_{\xx}u,\nabla_{\xx}\varphi\rangle\,\dm\mu$, so a maximizer satisfies $-\nabla_{\xx}\!\cdot(\omega\nabla_{\xx}u^\star)=q-p_{\mtheta}$ weakly. Testing this equation with $u^\star$ gives $\int u^\star(q-p_{\mtheta})\,\dm\xx=\int\|\nabla_{\xx}u^\star\|^2\dm\mu$, hence $J(u^\star)=\frac12\int\|\nabla_{\xx}u^\star\|^2\dm\mu=\frac12\|q-p_{\mtheta}\|^2_{\dot{H}^{-1}(\mu)}$, using the dual definition $\|q-p_{\mtheta}\|_{\dot{H}^{-1}(\mu)}=\sup\{\int u(q-p_{\mtheta})\,\dm\xx:\int\|\nabla_{\xx}u\|^2\dm\mu\le1\}$.

\section{Protocol for the controlled 2D study}
\label{app:toy}
The released script \texttt{experiments/toy8\_evolution\_figure.py} uses a mixture of eight Gaussians on a circle (radius 2, standard deviation 0.16), three-hidden-layer SiLU MLP potential and generator networks (width 128), batch size 512, Adam with learning rate $2\times10^{-4}$ for both networks, and two critic updates per generator update. The paired visualization in \figref{fig:toy8-comparison} uses seed 0 and 30K generator updates for each method. In that rerun, \NextGAN covers all eight modes with effective mode count 7.995 and captured fraction 0.928; the matched GAN baseline has effective mode count 2.770 and captured fraction 0.933. The raw independent-pair cosine is not a field-recovery metric because the population target is a conditional average rather than an individual displacement. If reporting direction accuracy, we first compute the numerical conditional field on a 2D grid.

\section{Regularity of the field identity}
For fixed conditioning variables and a differentiable potential, Eq.~\eqref{eq:line-integral} is the fundamental theorem of calculus applied to the straight path in Eq.~\eqref{eq:corridor}. The identity concerns the population field-regression objective and requires the same conditioning variables along each path. It does not require a pointwise optimal-transport coupling; the real--fake pairing is the one sampled by the training procedure.

\section{Additional Discussion}
\label{app:additional-discussions}
Throughout this appendix, we fix a condition $\cc$ and suppress it in the notation; $P=P_{\mtheta}(\cdot\mid\cc)$ and $Q=Q(\cdot\mid\cc)$. We use $\alpha$ for interpolation time and $t$ for distributional evolution. This appendix separates derivation from verification. We first develop the geometric construction from quadratic transport to the practical \NextGAN objective and clarify its scope. \appref{app:proofs} then states the formal assumptions and proves the conditional-regression, variational, differentiation, and minibatch results.

\subsection{Wasserstein Gradient Flows of Energy Functionals}
\label{app:wgf}
We first place the NextGAN construction in the general Wasserstein-gradient-flow (WGF) framework~\citep{han2026wflow}. Let $P_t$ be an evolving probability measure on the comparison space and let $Q$ be a fixed target measure. For an energy functional $\cF(P;Q)$ with first variation $\delta\cF/\delta P$, its formal WGF is
\begin{equation}
\partial_t P_t
=\nabla_{\xx}\!\cdot\!\left(P_t\nabla_{\xx}\frac{\delta\cF}{\delta P}(P_t;Q)\right),
\label{eq:app-wgf-density}
\end{equation}
or, equivalently, the continuity equation
\begin{equation}
\partial_tP_t+\nabla_{\xx}\!\cdot(P_t\vv_t)=0,
\qquad
\vv_t(\xx)=-\nabla_{\xx}\frac{\delta\cF}{\delta P}(P_t;Q)(\xx).
\label{eq:app-wgf-continuity}
\end{equation}
When the differentiations are justified and there is no boundary flux, the energy decreases along the flow:
\begin{equation}
\frac{\dm}{\dm t}\cF(P_t;Q)
=-\int\left\|\nabla_{\xx}\frac{\delta\cF}{\delta P}(P_t;Q)(\xx)\right\|^2P_t(\dm\xx)
\leq 0.
\label{eq:app-wgf-dissipation}
\end{equation}
Thus an energy specifies a distributional velocity field, while a numerical algorithm only approximates the corresponding flow.

\subsubsection{The NextGAN energy as a distributional functional}
For a fixed nonnegative path-occupation measure $\mu$, define the population NextGAN energy of a generated feature distribution $P$ relative to a real feature distribution $Q$ by
\begin{equation}
\cF_{\mathrm{cor}}^{\mu}(P;Q)
=\sup_{u\in\cH_\mu}
\left\{
\int u\,\dm Q-\int u\,\dm P
-\frac12\int\|\nabla_{\xx}u(\xx)\|^2\,\mu(\dm\xx)
\right\},
\label{eq:app-cor-energy}
\end{equation}
where $\cH_\mu$ is the finite-energy space modulo the kernel of the seminorm $u\mapsto(\int\|\nabla_{\xx}u\|^2\dm\mu)^{1/2}$. Under the assumptions in \appref{app:proofs}, let $u_P^\star$ be an optimizer of Eq.~\eqref{eq:app-cor-energy}. The envelope theorem gives the first variation with respect to the generated measure:
\begin{equation}
\frac{\delta\cF_{\mathrm{cor}}^{\mu}}{\delta P}(P;Q)(\xx)
=-u_P^\star(\xx)+\mathrm{const},
\label{eq:app-cor-first-variation}
\end{equation}
where the additive constant is immaterial because probability-preserving perturbations have zero total mass. Substituting Eq.~\eqref{eq:app-cor-first-variation} into the WGF velocity in Eq.~\eqref{eq:app-wgf-continuity} yields
\begin{equation}
\vv_t(\xx)=\nabla_{\xx}u_{P_t}^\star(\xx).
\label{eq:app-cor-wgf-velocity}
\end{equation}
Therefore, at the population level and for fixed $\mu$, the NextGAN discriminator potential is the scalar potential whose gradient gives the WGF velocity of the generated comparison-space distribution. The generator loss $-\E_{\xx\sim P_t}[u_{P_t}^\star(\xx)]$ moves generated particles in the same direction, since gradient descent on the particle energy $-u_{P_t}^\star$ produces $\dot{\xx}=\nabla_{\xx}u_{P_t}^\star(\xx)$.

For fixed $\mu$, Eq.~\eqref{eq:app-wgf-dissipation} specializes to
\begin{equation}
\frac{\dm}{\dm t}\cF_{\mathrm{cor}}^{\mu}(P_t;Q)
=-\int\|\nabla_{\xx}u_{P_t}^\star(\xx)\|^2P_t(\dm\xx).
\label{eq:app-cor-dissipation}
\end{equation}
The energy term in Eq.~\eqref{eq:app-cor-energy} is instead integrated against the path measure $\mu$. These two measures coincide only under an additional identification such as $\mu=P_t$; if $\mu$ changes with $t$, the geometry-variation term derived in \appref{app:proofs} must also be included. The WGF interpretation is exact for the population \NextGAN functional under the stated assumptions. It does not extend automatically to alternative logistic or WGAN objectives, or to finite-width alternating neural optimization.

\subsection{Quadratic Wasserstein Geometry}
\label{app:ot-geometry}

\subsubsection{Static and dynamic formulations}
The Kantorovich definition is
\begin{equation}
W_2^2(P,Q)=\inf_{\pi\in\Pi(P,Q)}\E_{(\xx_{\mathrm{p}},\xx_{\mathrm{r}})\sim\pi}\|\xx_{\mathrm{r}}-\xx_{\mathrm{p}}\|^2.
\label{eq:app-w2-static}
\end{equation}
Under the standard assumptions of the Benamou--Brenier formulation, this is equivalently
\begin{equation}
\frac12W_2^2(P,Q)
=\inf_{\rho_t,\vv_t}\left\{
\int_0^1\!\int\frac12\rho_t(\xx)\|\vv_t(\xx)\|^2\,\dm\xx\,\dm t
\;\middle|\;
\begin{aligned}
&\partial_t\rho_t+\nabla_{\xx}\!\cdot(\rho_t\vv_t)=0,\\[-1pt]
&\rho_{t=0}=P,\qquad \rho_{t=1}=Q
\end{aligned}\right\}.
\label{eq:app-bb}
\end{equation}
The objective is the kinetic energy of the transporting velocity, and the constraints enforce mass conservation and the endpoint marginals~\citep{benamou2000computational}. Introducing a multiplier $\Phi(t,\xx)$ for the continuity equation and integrating both the time-derivative and divergence terms by parts, the Lagrangian (up to fixed-endpoint boundary terms) becomes
\begin{equation}
\cL(\rho,\vv,\Phi)
=\int_0^1\!\int\rho_t(\xx)\left[
\frac12\|\vv_t(\xx)\|^2-\partial_t\Phi(t,\xx)-\nabla_{\xx}\Phi(t,\xx)^\top\vv_t(\xx)
\right]\dm\xx\,\dm t.
\label{eq:app-integrated-lagrangian}
\end{equation}
Stationarity with respect to $\vv_t$ gives $\rho_t(\vv_t-\nabla_{\xx}\Phi)=0$, hence
\begin{equation}
\vv_t(\xx)=\nabla_{\xx}\Phi(t,\xx)\qquad\text{on the support of }\rho_t.
\label{eq:app-conservative-ot}
\end{equation}
Substituting this relation into Eq.~\eqref{eq:app-integrated-lagrangian} and taking the first variation with respect to $\rho_t$ gives the Hamilton--Jacobi equation
\begin{equation}
\partial_t\Phi(t,\xx)+\frac12\|\nabla_{\xx}\Phi(t,\xx)\|^2=0
\qquad\text{where }\rho_t(\xx)>0.
\label{eq:app-hj}
\end{equation}
If $\Phi$ is twice continuously differentiable, equality of mixed partials makes the corresponding velocity locally irrotational in any dimension:
\begin{equation}
\partial_{x_i}(\vv_t)_j-\partial_{x_j}(\vv_t)_i=0,\qquad i,j=1,\ldots,d.
\label{eq:app-ot-irrotational}
\end{equation}
Thus, the conservative velocity and the Hamilton--Jacobi potential follow from the dynamic OT first-order conditions. The potential is generally time-dependent; the curl-free statement is specific to the OT optimum and is not imposed on an arbitrary flow-matching field.

\subsubsection{Kantorovich dual and the OT boundary}
For completeness, the Kantorovich dual of the quadratic cost is
\begin{equation}
\frac12 W_2^2(P,Q)
=\sup_{(\varphi,\psi)}
\left\{\int\varphi\,\dm P+\int\psi\,\dm Q:\;
\varphi(\xx_{\mathrm{p}})+\psi(\xx_{\mathrm{r}})\leq \frac12\|\xx_{\mathrm{r}}-\xx_{\mathrm{p}}\|^2\right\}.
\label{eq:app-kantorovich-dual}
\end{equation}
The constraint in Eq.~\eqref{eq:app-kantorovich-dual} is imposed for every endpoint pair and the coupling is optimized implicitly by the duality theorem. In contrast, Eq.~\eqref{eq:app-value} uses one static potential, samples the fixed product coupling, and penalizes its gradient only on the resulting straight-path occupation measure. Consequently, replacing the OT coupling by $P\times Q$ changes the objective before any neural approximation is introduced.

\subsection{Flow Matching for a Prescribed Coupling}
\label{app:flow-matching}
\subsubsection{Displacement paths}
For any fixed coupling $\pi\in\Pi(P,Q)$, draw $(\xx_{\mathrm{p}},\xx_{\mathrm{r}})\sim\pi$ and define the straight path
\begin{equation}
\xx_\alpha=(1-\alpha)\xx_{\mathrm{p}}+\alpha\xx_{\mathrm{r}}.
\label{eq:app-path}
\end{equation}
The induced law is $\rho_\alpha=\operatorname{Law}(\mX_\alpha)$. If $\pi=\pi^\star$ is an optimal coupling, this law is the constant-speed $W_2$ geodesic and its Eulerian velocity satisfies
\begin{equation}
\vv_\alpha^\star(\xx)=\E_{\pi^\star}[\mX_{\mathrm{r}}-\mX_{\mathrm{p}}\mid\mX_\alpha=\xx,\alpha],
\qquad
\int_0^1\E_{\xx\sim\rho_\alpha}\|\vv_\alpha^\star(\xx)\|^2\dm\alpha=W_2^2(P,Q).
\label{eq:app-action}
\end{equation}
For an arbitrary coupling, optimality of $W_2$ gives
\begin{equation}
W_2^2(P,Q)\leq \E_{\pi}\|\xx_{\mathrm{r}}-\xx_{\mathrm{p}}\|^2,
\label{eq:app-coupling-gap}
\end{equation}
with equality only when the chosen coupling is optimal (up to non-uniqueness). This inequality records the distinction between optimizing the OT coupling and prescribing a coupling for flow matching. Under the product coupling used by NextGAN, the straight-path law serves as a tractable reference path distribution and need not coincide with the Wasserstein geodesic.

\subsubsection{Conditional flow-matching regression}
For a general prescribed coupling, define the conditional velocity $\vv_\alpha^\pi(\xx)=\E_\pi[\mX_{\mathrm{r}}-\mX_{\mathrm{p}}\mid\mX_\alpha=\xx,\alpha]$. For any measurable field $\ww_\alpha$,
\begin{equation}
\E\|\ww_\alpha(\xx_\alpha)-(\xx_{\mathrm{r}}-\xx_{\mathrm{p}})\|^2
=\E\|\ww_\alpha(\xx_\alpha)-\vv_\alpha^\pi(\xx_\alpha)\|^2
+\E\|\vv_\alpha^\pi(\xx_\alpha)-(\xx_{\mathrm{r}}-\xx_{\mathrm{p}})\|^2,
\label{eq:app-conditional-decomp}
\end{equation}
Therefore, the flow-matching regression is minimized by $\ww_\alpha=\vv_\alpha^\pi$ almost everywhere. This statement is valid for any chosen coupling and does not require that the coupling be optimal.

\subsection{From Flow Matching to the NextGAN Objective}
\label{app:nextgan-objective}
At the fixed condition $\cc$, $\cL_{\mathrm{field}}(u)$ denotes the conditional version of Eq.~\eqref{eq:field-regression} for a generic potential $u$. Under the product coupling $\pi=P\times Q$ and the conservative potential field described in \appref{app:wgf}, the NextGAN flow-matching regression is
\begin{equation}
2\cL_{\mathrm{field}}(u)=\E\|\nabla_{\xx}u(\xx_\alpha)-(\xx_{\mathrm{r}}-\xx_{\mathrm{p}})\|^2.
\label{eq:app-fm}
\end{equation}

\subsubsection{Algebraic reduction}
Using $u(\xx_{\mathrm{r}})-u(\xx_{\mathrm{p}})=\int_0^1\nabla_{\xx}u(\xx_\alpha)^\top(\xx_{\mathrm{r}}-\xx_{\mathrm{p}})\,\dm\alpha$ and uniform $\alpha$, expansion of Eq.~\eqref{eq:app-fm} gives
\begin{equation}
2\cL_{\mathrm{field}}(u)=\E\|\nabla_{\xx}u(\xx_\alpha)\|^2-2\E[u(\xx_{\mathrm{r}})-u(\xx_{\mathrm{p}})]+C,
\qquad C=\E\|\xx_{\mathrm{r}}-\xx_{\mathrm{p}}\|^2.
\label{eq:appendix-identity}
\end{equation}
The constant $C$ is independent of $u$. Therefore, minimizing Eq.~\eqref{eq:appendix-identity} over the conservative field is equivalent to maximizing
\begin{equation}
\cV(u)=\E[u(\xx_{\mathrm{r}})-u(\xx_{\mathrm{p}})]-\frac12\E\|\nabla_{\xx}u(\xx_\alpha)\|^2.
\label{eq:app-value}
\end{equation}
Eq.~\eqref{eq:app-value} is the NextGAN objective. It is a distinct adversarial functional motivated by the quadratic-transport geometry and instantiated through flow matching, rather than an exact evaluation of $W_2$.

\subsection{Scope and Relation to Optimal Transport}
\label{app:boundaries}
Exact quadratic OT optimizes the endpoint coupling and a density path under the continuity equation, with a time-dependent cost-conjugate potential~\citep{villani2009optimal,benamou2000computational}. NextGAN instead fixes $P\times Q$, prescribes straight paths, and uses a static potential. These restrictions explain why Eq.~\eqref{eq:app-value} should be interpreted as a $W_2$-motivated adversarial objective rather than as an exact Wasserstein distance.

The formal variational characterization, well-posedness conditions, and implementation-level gradient rules are stated and proved in \appref{app:proofs}.

\section{Formal Statements and Proofs}
\label{app:proofs}
\label{app:identity}
This section gives the formal results used by the construction in \appref{app:additional-discussions}. The results concern population objectives and their Monte Carlo estimators; they do not constitute a convergence theorem for finite-width neural-network alternating optimization.

\subsection{Assumptions and endpoint conventions}
\begin{assumption}[NextGAN regularity and endpoint convention]
\label{assump:nextgan-regularity}
The comparison space is $\R^d$, and $P,Q\in\cP_2(\R^d)$, the set of probability measures with finite second moment. We use $\xx_{\mathrm{p}}\sim P$ for a generated comparison vector and $\xx_{\mathrm{r}}\sim Q$ for a real comparison vector. The product coupling $P\times Q$ is used in the practical objective, while $\pi^\star$ denotes an optimizer of the quadratic transport problem when it exists. Expectations involving a path also integrate an independent $\alpha\sim\mathcal{U}(0,1)$. The functions and vector fields appearing below are measurable and square-integrable under the stated path laws, and differentiation under an expectation is used only when the corresponding integrability and regularity conditions hold.
\end{assumption}

The regularity conditions above are sufficient for the displayed formal calculations; the neural-network implementation is a finite-dimensional approximation and does not satisfy them automatically.

\begin{proposition}[Conditional flow-matching decomposition]
\label{prop:app-cfm}
Let $(\xx_{\mathrm{p}},\xx_{\mathrm{r}})\sim\pi$ have finite second moment, let $\xx_\alpha=(1-\alpha)\xx_{\mathrm{p}}+\alpha\xx_{\mathrm{r}}$, and define $\vv_\alpha^\pi(\xx)=\E[\mX_{\mathrm{r}}-\mX_{\mathrm{p}}\mid\mX_\alpha=\xx,\alpha]$. For every square-integrable measurable field $\ww_\alpha$,
\begin{equation}
\E\|\ww_\alpha(\xx_\alpha)-(\xx_{\mathrm{r}}-\xx_{\mathrm{p}})\|^2
=\E\|\ww_\alpha(\xx_\alpha)-\vv_\alpha^\pi(\xx_\alpha)\|^2
+\E\|\vv_\alpha^\pi(\xx_\alpha)-(\xx_{\mathrm{r}}-\xx_{\mathrm{p}})\|^2.
\label{eq:app-cfm-proof-result}
\end{equation}
Thus $\vv_\alpha^\pi$ is the unique minimizer up to $\rho_\alpha$-null sets for almost every $\alpha$, where $\rho_\alpha=\operatorname{Law}(\mX_\alpha)$.
\end{proposition}

\begin{proof}
Write $\aa=\ww_\alpha(\xx_\alpha)-\vv_\alpha^\pi(\xx_\alpha)$ and $\bb=\vv_\alpha^\pi(\xx_\alpha)-(\xx_{\mathrm{r}}-\xx_{\mathrm{p}})$. Expanding $\|\aa+\bb\|^2$ leaves the cross term $2\E[\aa^\top\bb]$. Conditioning on $(\xx_\alpha,\alpha)$ gives
\[
\E[\aa^\top\bb\mid\xx_\alpha,\alpha]
=\aa^\top\E[\vv_\alpha^\pi(\xx_\alpha)-(\xx_{\mathrm{r}}-\xx_{\mathrm{p}})\mid\xx_\alpha,\alpha]=0,
\]
by the definition of the conditional expectation. Taking expectations proves Eq.~\eqref{eq:app-cfm-proof-result}; the first term is nonnegative and vanishes exactly when $\ww_\alpha(\xx_\alpha)=\vv_\alpha^\pi(\xx_\alpha)$ almost everywhere.
\end{proof}

\begin{proposition}[Algebraic reduction to the NextGAN value]
\label{prop:app-algebra}
Assume $u\in C^1$ and that the terms below are integrable. Under any coupling with straight paths,
\begin{equation}
\E\|\nabla_{\xx}u(\xx_\alpha)-(\xx_{\mathrm{r}}-\xx_{\mathrm{p}})\|^2
=\E\|\nabla_{\xx}u(\xx_\alpha)\|^2
-2\E[u(\xx_{\mathrm{r}})-u(\xx_{\mathrm{p}})]
+\E\|\xx_{\mathrm{r}}-\xx_{\mathrm{p}}\|^2.
\label{eq:app-algebra-proof-result}
\end{equation}
Therefore minimizing the flow-matching regression over the conservative potential class used by the method is equivalent to maximizing Eq.~\eqref{eq:app-value}.
\end{proposition}

\begin{proof}
Expand the square and use the fundamental theorem of calculus along each line segment:
\[
u(\xx_{\mathrm{r}})-u(\xx_{\mathrm{p}})=\int_0^1\nabla_{\xx}u(\xx_\alpha)^\top(\xx_{\mathrm{r}}-\xx_{\mathrm{p}})\,\dm\alpha.
\]
Since $\alpha$ is uniform, $\E[\nabla_{\xx}u(\xx_\alpha)^\top(\xx_{\mathrm{r}}-\xx_{\mathrm{p}})]=\E[u(\xx_{\mathrm{r}})-u(\xx_{\mathrm{p}})]$. Substitution into the expanded square gives Eq.~\eqref{eq:app-algebra-proof-result}; the final displacement term is independent of $u$.
\end{proof}

\subsection{Population variational characterization}
\label{app:population-variational}
For a fixed condition, let $p,q$ be the generated and real feature densities and let $\mu$ be the occupation measure of the sampled paths, with density $\omega$. Assume that $r=q-p$ defines a bounded linear functional on the finite-energy space and annihilates the kernel of its energy seminorm. The population objective is
\begin{equation}
\cV_\mu(p,u)
=\int u(\xx)(q(\xx)-p(\xx))\,\dm\xx
-\frac12\int\omega(\xx)\|\nabla_{\xx}u(\xx)\|^2\,\dm\xx.
\label{eq:app-population}
\end{equation}
For a perturbation $u+\varepsilon v$, its first variation is
\begin{equation}
\left.\frac{\dm}{\dm\varepsilon}\cV_\mu(p,u+\varepsilon v)\right|_{\varepsilon=0}
=\int v(\xx)(q(\xx)-p(\xx))\,\dm\xx
-\int\omega(\xx)\nabla_{\xx}u(\xx)\!\cdot\!\nabla_{\xx}v(\xx)\,\dm\xx.
\label{eq:app-first-variation}
\end{equation}
Stationarity for every admissible test function $v$ gives the weak Euler--Lagrange equation
\begin{equation}
-\nabla_{\xx}\!\cdot(\omega\nabla_{\xx}u^\star)=q-p.
\label{eq:app-poisson}
\end{equation}
To state the associated dual value, define the energy form
\begin{equation}
a_\mu(u_1,u_2)=\int \omega(\xx)\nabla_{\xx}u_1(\xx)^\top\nabla_{\xx}u_2(\xx)\,\dm\xx
\label{eq:app-energy-form}
\end{equation}
and assume that it is coercive after quotienting out its kernel (or after fixing a gauge such as zero mean). Define the dual seminorm
\begin{equation}
\|r\|_{\dot{H}^{-1}(\mu)}
=\sup_{a_\mu(v,v)>0}\frac{\int r(\xx)v(\xx)\,\dm\xx}{a_\mu(v,v)^{1/2}}.
\label{eq:app-dual-norm}
\end{equation}

\begin{theorem}[Weighted Poisson characterization of the NextGAN value]
\label{thm:app-poisson}
Under the assumptions above, there is a unique equivalence class $u^\star$ satisfying
\begin{equation}
a_\mu(u^\star,v)=\int r(\xx)v(\xx)\,\dm\xx
\qquad\text{for every admissible }v,
\label{eq:app-weak-poisson}
\end{equation}
which is the weak form of $-\nabla_{\xx}\!\cdot(\omega\nabla_{\xx}u^\star)=r$. Moreover,
\begin{equation}
\sup_u\left\{\int u r\,\dm\xx-\frac12a_\mu(u,u)\right\}
=\frac12a_\mu(u^\star,u^\star)
=\frac12\|r\|_{\dot{H}^{-1}(\mu)}^2.
\label{eq:app-hminusone}
\end{equation}
\end{theorem}

\begin{proof}
Coercivity and boundedness allow the Riesz representation of the functional $v\mapsto\int rv\,\dm\xx$ in the energy inner product, yielding Eq.~\eqref{eq:app-weak-poisson}. For any admissible $u$, write $u=u^\star+w$. Using Eq.~\eqref{eq:app-weak-poisson},
\[
\int ur\,\dm\xx-\frac12a_\mu(u,u)
=\frac12a_\mu(u^\star,u^\star)-\frac12a_\mu(w,w)
\leq \frac12a_\mu(u^\star,u^\star).
\]
Equality holds exactly for $w$ in the kernel, proving the first equality in Eq.~\eqref{eq:app-hminusone}. Finally, the Riesz representer identity gives
\[
\|r\|_{\dot{H}^{-1}(\mu)}^2=a_\mu(u^\star,u^\star),
\]
which proves the second equality.
\end{proof}

The theorem is a population statement for a fixed path geometry. It does not imply that alternating stochastic gradient updates find $u^\star$, nor that a finite neural discriminator realizes the whole energy space. It also does not identify the value with $W_2^2(P,Q)$, because the product coupling, straight paths, static potential, and occupation measure $\mu$ are fixed by the NextGAN construction.

\subsection{Variation of the path geometry}
\label{app:geometry-variation}
Let $s$ index endpoint densities $p_s,q_s$ and occupation measures $\mu_s$ with densities $\omega_s$. Write $r_s=q_s-p_s$, and suppose that $u_s$ is an exact stationary maximizer of
\begin{equation}
\cV(s,u)=\int u(\xx)r_s(\xx)\,\dm\xx
-\frac12\int\omega_s(\xx)\|\nabla_{\xx}u(\xx)\|^2\,\dm\xx.
\label{eq:app-adaptive-value}
\end{equation}
Under differentiability and an envelope-theorem interchange of derivative and integral,
\begin{equation}
\frac{\dm}{\dm s}\cV(s,u_s)
=\int u_s(\xx)\,\dot r_s(\xx)\,\dm\xx
-\frac12\int \dot\omega_s(\xx)\|\nabla_{\xx}u_s(\xx)\|^2\,\dm\xx.
\label{eq:app-geometry-variation}
\end{equation}
The terms involving $\dot u_s$ vanish because $u_s$ is stationary in the discriminator variable. The second term is the geometry-variation contribution; its sign is determined by $\dot\omega_s$ and therefore is not fixed in general. Eq.~\eqref{eq:app-geometry-variation} is why the fixed-geometry interpretation cannot be transferred unchanged to an adaptive path distribution.

\subsection{Differentiation through a frozen representer}
For a raw generated sample $\bar{\xx}_{\mathrm{p}}=\mg_{\mtheta}(\zz,\cc)$ and its comparison-space representation $\xx_{\mathrm{p}}=\mf(\bar{\xx}_{\mathrm{p}},\cc)$, the generator-side chain rule is
\begin{equation}
\nabla_{\mtheta}u_{\mphi}(\xx_{\mathrm{p}},\cc)
=J_{\mg_{\mtheta}}(\zz,\cc)^\top
J_{\mf}(\bar{\xx}_{\mathrm{p}},\cc)^\top
\nabla_{\xx}u_{\mphi}(\xx_{\mathrm{p}},\cc).
\label{eq:app-frozen-chain-rule}
\end{equation}
Freezing $\mf$ means that its parameters receive no update; it does not remove the input Jacobian $J_{\mf}$ needed to propagate the discriminator signal to $\mg_{\mtheta}$. With multiple independent representation fields and scalar weights $w_\ell$ (unit weights for the summed head losses), linearity gives
\begin{equation}
\nabla_{\mtheta}\sum_{\ell=1}^Lw_\ell
u_{\mphi_\ell}\!\left(\mf_\ell(\mg_{\mtheta}(\zz,\cc),\cc),\cc\right)
=\sum_{\ell=1}^Lw_\ell
J_{\mg_{\mtheta}}^\top J_{\mf_\ell}^\top
\nabla_{\xx_\ell}u_{\mphi_\ell}.
\label{eq:app-multifield-chain-rule}
\end{equation}
During the discriminator update, generated and real endpoint vectors are detached from $\mg_{\mtheta}$ and the representation-map parameters; during the generator update, discriminator parameters are held fixed while the chain rule above is retained.

\subsection{Finite-sample alternating objectives}
For a minibatch of size $B$, independently draw $\cc_i\sim p_{\mathrm{data}}(\cc)$ and endpoint pairs $(\xx_{\mathrm{p}}^i,\xx_{\mathrm{r}}^i)\sim P_{\mtheta}(\cdot\mid\cc_i)\times Q(\cdot\mid\cc_i)$. Independently draw $\alpha_i\sim\mathcal{U}(0,1)$ and set $\xx_{\alpha_i}^i=(1-\alpha_i)\xx_{\mathrm{p}}^i+\alpha_i\xx_{\mathrm{r}}^i$. The Monte Carlo discriminator loss corresponding to Eq.~\eqref{eq:app-value} is
\begin{equation}
\widehat{\cL}_{\mathrm{D}}^{\mathrm{adv}}
=\frac1B\sum_{i=1}^B\left[
u_{\mphi}(\xx_{\mathrm{p}}^i,\cc_i)-u_{\mphi}(\xx_{\mathrm{r}}^i,\cc_i)
+\frac12\left\|\nabla_{\xx}u_{\mphi}(\xx_{\alpha_i}^i,\cc_i)\right\|^2\right].
\label{eq:app-empirical-d}
\end{equation}
For this update, the endpoint vectors are treated as constants with respect to $\mg_{\mtheta}$ and the representation-map parameters, while $\xx_{\alpha_i}^i$ remains a differentiable input of $u_{\mphi}$ so that the field energy can be evaluated. With discriminator parameters held fixed at $\mphi$, the generator loss is
\begin{equation}
\widehat{\cL}_{\mathrm{gen}}
=-\frac1B\sum_{i=1}^B u_{\mphi}(\xx_{\mathrm{p}}^i,\cc_i).
\label{eq:app-empirical-g}
\end{equation}
For parameters fixed independently of the sampled minibatch, Eqs.~\eqref{eq:app-empirical-d}--\eqref{eq:app-empirical-g} are unbiased estimators of their population counterparts before the nonlinear optimization update. The gradient of Eq.~\eqref{eq:app-empirical-g} is taken through $\xx_{\mathrm{p}}^i=\mf(\mg_{\mtheta}(\zz_i,\cc_i),\cc_i)$, whereas no gradient is propagated into $\mphi$ or the frozen representation-map parameters.

\begin{proposition}[Unbiasedness of the minibatch objectives]
\label{prop:app-unbiased}
Under \assumpref{assump:nextgan-regularity}, fix $\mtheta,\mphi$ independently of the minibatch. If the endpoint pairs and interpolation times in Eq.~\eqref{eq:app-empirical-d} are independent and identically distributed, then
\begin{equation}
\E\left[\widehat{\cL}_{\mathrm{D}}^{\mathrm{adv}}\right]
=\cL_{\mathrm{D}}^{\mathrm{adv}},
\qquad
\E\left[\widehat{\cL}_{\mathrm{gen}}\right]
=\cL_{\mathrm{gen}}.
\label{eq:app-unbiased}
\end{equation}
\end{proposition}

\begin{proof}
Each summand in Eqs.~\eqref{eq:app-empirical-d} and~\eqref{eq:app-empirical-g} has the same expectation as the corresponding population integrand by the sampling rule and \assumpref{assump:nextgan-regularity}. Linearity of expectation and the factor $1/B$ then give Eq.~\eqref{eq:app-unbiased}. The statement concerns the objective estimator at fixed parameters; it does not assert unbiasedness of a finite-step optimizer after parameter updates.
\end{proof}

\section{Convergence of the NextGAN Particle Dynamics}
\label{app:convergence}
The preceding sections identify the population NextGAN energy and its WGF velocity. We now analyze the comparison-space particle dynamics induced by that velocity, following the particle-to-continuum distinction used in distributional WGF analyses. This result is an idealized dynamics statement: the critic is represented by a velocity oracle up to an explicit approximation error, and the conclusion is not a convergence theorem for finite-width neural-network alternating optimization.

\subsection{NextGAN velocity and empirical dynamics}
Fix a nonnegative occupation measure $\mu$ and a real comparison-space distribution $Q$. For a generated comparison-space distribution $P$, let $u_{P,Q}^{\mu}$ be a maximizer of Eq.~\eqref{eq:app-cor-energy} and define
\begin{equation}
\vv_{P,Q}^{\mu}(\xx):=\nabla_{\xx}u_{P,Q}^{\mu}(\xx).
\label{eq:app-nextgan-velocity}
\end{equation}
By Eq.~\eqref{eq:app-cor-first-variation}, this is the formal WGF velocity of the NextGAN energy with respect to its generated marginal when $\mu$ is fixed. The corresponding continuum dynamics are
\begin{equation}
\partial_tP_t+\nabla_{\xx}\!\cdot\!\left(P_t\vv_{P_t,Q}^{\mu}\right)=0,
\qquad P_{t=0}=P_0.
\label{eq:app-nextgan-pde}
\end{equation}

Given generated comparison-space particles $\{\xx_i^{(0)}\}_{i=1}^N$, a real empirical measure $\widehat Q_M=M^{-1}\sum_{j=1}^M\delta_{\yy_j}$, and step size $\eta>0$, define
\begin{equation}
\widehat P_k=\frac1N\sum_{i=1}^N\delta_{\xx_i^{(k)}},
\qquad
\xx_i^{(k+1)}=\xx_i^{(k)}+\eta\widetilde{\vv}_k(\xx_i^{(k)}),
\label{eq:app-nextgan-particle-update}
\end{equation}
where $\widetilde{\vv}_k$ is the velocity produced by the learned discriminator. The velocity residual appearing below is evaluated directly at the current particles, without introducing an additional error variable.
Let $\widehat P_t^{\eta}$ denote the empirical measure obtained by linearly interpolating the particles between $t_k=k\eta$ and $t_{k+1}$.

\subsection{Regularity assumption}
\begin{assumption}[NextGAN velocity regularity]
\label{assump:nextgan-velocity}
Write $m_2(R):=(\int\|\xx\|_2^2\,\dm R(\xx))^{1/2}$. There exist finite constants $L_0,L_x,L_P,L_Q$ such that, for all comparison-space measures $P,\widetilde P,Q,\widetilde Q$ in the relevant bounded-moment class:
\begin{align}
\|\vv_{P,Q}^{\mu}\|_{L^2(P)}
&\leq L_0\bigl(1+m_2(P)+m_2(Q)\bigr),\label{eq:app-cv-growth}\\
\|\vv_{P,Q}^{\mu}(\xx)-\vv_{P,Q}^{\mu}(\xx')\|
&\leq L_x\|\xx-\xx'\|,\label{eq:app-cv-space-lip}\\
\|\vv_{P,Q}^{\mu}-\vv_{\widetilde P,Q}^{\mu}\|_{L^2(\widetilde P)}
&\leq L_PW_2(P,\widetilde P),\label{eq:app-cv-source-lip}\\
\|\vv_{P,Q}^{\mu}-\vv_{P,\widetilde Q}^{\mu}\|_{L^2(P)}
&\leq L_QW_2(Q,\widetilde Q).\label{eq:app-cv-target-lip}
\end{align}
The initial comparison-space particles are i.i.d. from $P_0$, the real particles are i.i.d. from $Q$, and the velocity residuals evaluated at the current particles are square-integrable. The fixed measure $\mu$ and the admissible comparison-space domain are shared by the continuum and empirical dynamics.
\end{assumption}

\assumpref{assump:nextgan-velocity} is the direct analogue of the velocity regularity conditions used for particle-dynamics convergence. For NextGAN, it is a condition to be checked for the selected representation geometry and discriminator class; spectral normalization and bounded comparison-space domains can support such a check, but do not by themselves imply all four inequalities.

\subsection{Continuum and particle convergence}
\begin{theorem}[Convergence of NextGAN-induced particle dynamics]
\label{thm:nextgan-convergence}
Assume \assumpref{assump:nextgan-velocity} and let $T>0$. Set $L=L_x+L_P$ (with $(e^{LT}-1)/L$ interpreted as $T$ when $L=0$) and $K_T=\lceil T/\eta\rceil$. Then Eq.~\eqref{eq:app-nextgan-pde} has a unique weak solution $P\in C([0,T];\cP_2(\R^d))$. There is a constant $C_T<\infty$, depending only on $T$, the constants in \assumpref{assump:nextgan-velocity}, and the shared moment bound, such that
\begin{equation}
\begin{aligned}
\sup_{t\in[0,T]}W_2\!\left(\widehat P_t^{\eta},P_t\right)
&\leq e^{LT}W_2(\widehat P_0,P_0)
+\frac{L_Q}{L}\bigl(e^{LT}-1\bigr)W_2(\widehat Q_M,Q) \\
&\quad +C_T\eta
+e^{LT}\eta\sum_{k=0}^{K_T-1}
\left(\frac1N\sum_{i=1}^N
\left\|\widetilde{\vv}_k(\xx_i^{(k)})-\vv_{\widehat P_k,\widehat Q_M}^{\mu}(\xx_i^{(k)})\right\|^2\right)^{1/2}.
\end{aligned}
\label{eq:app-nextgan-convergence-bound}
\end{equation}
Consequently, if $N,M\to\infty$, $\eta\to0$, and the explicitly displayed weighted residual sum converges to zero, then
\begin{equation}
\sup_{t\in[0,T]}W_2\!\left(\widehat P_t^{\eta},P_t\right)\longrightarrow0
\qquad\text{almost surely}.
\label{eq:app-nextgan-convergence-limit}
\end{equation}
\end{theorem}

\begin{proof}
We separate the proof into the continuum stability, the empirical particle system, the Euler error, and the learned-velocity error.

\emph{Continuum stability.} Let $P_t$ and $\widetilde P_t$ solve Eq.~\eqref{eq:app-nextgan-pde} with initial measures $P_0,\widetilde P_0$ and target measures $Q,\widetilde Q$. Couple their characteristic curves using an optimal initial coupling. The spatial and source-measure bounds in \assumpref{assump:nextgan-velocity}, followed by Cauchy--Schwarz, give for the induced distance $D(t)$
\[
\frac{\dm}{\dm t}D(t)
\leq L D(t)+L_QW_2(Q,\widetilde Q).
\]
Gronwall's inequality yields
\begin{equation}
W_2(P_t,\widetilde P_t)
\leq e^{Lt}W_2(P_0,\widetilde P_0)
+\frac{L_Q}{L}\bigl(e^{Lt}-1\bigr)W_2(Q,\widetilde Q).
\label{eq:app-nextgan-stability}
\end{equation}
Taking $\widetilde P_t$ to be the continuous-time empirical particle solution driven by $\vv_{\widehat P_t,\widehat Q_M}^{\mu}$ gives the first two terms of Eq.~\eqref{eq:app-nextgan-convergence-bound}.

\emph{Finite-dimensional well-posedness.} For $\mX=(\xx_1,\ldots,\xx_N)$, equip $(\R^d)^N$ with $\|\mX\|_{2,N}^2=N^{-1}\sum_i\|\xx_i\|^2$. Define
\[
\vv_N(\mX)=\left(\vv_{\widehat P_{\mX},\widehat Q_M}^{\mu}(\xx_1),\ldots,\vv_{\widehat P_{\mX},\widehat Q_M}^{\mu}(\xx_N)\right),
\]
where $\widehat P_{\mX}=N^{-1}\sum_i\delta_{\xx_i}$. Matching $\xx_i$ with $\xx_i'$ gives $W_2(\widehat P_{\mX},\widehat P_{\mX'})\leq\|\mX-\mX'\|_{2,N}$. Hence
\[
\|\vv_N(\mX)-\vv_N(\mX')\|_{2,N}
\leq(L_x+L_P)\|\mX-\mX'\|_{2,N}=L\|\mX-\mX'\|_{2,N}.
\]
The growth bound gives linear growth, so Picard--Lindelöf yields a unique global ODE solution for the continuous empirical particles. Its empirical measure is a weak solution of the corresponding empirical continuity equation.

\emph{Euler discretization.} Let $\bar{\mX}(t)$ denote the continuous empirical solution driven by the exact empirical velocity and let $\bar{\mX}^\eta(t)$ be its piecewise-linear Euler interpolation. The global Lipschitz bound and the growth estimate imply the standard Euler estimate
\begin{equation}
\sup_{t\in[0,T]}\|\bar{\mX}^\eta(t)-\bar{\mX}(t)\|_{2,N}\leq C_T\eta.
\label{eq:app-nextgan-euler-error}
\end{equation}
The coupling that matches particles with the same index gives the same bound for the associated empirical measures in $W_2$.

\emph{Learned-velocity error.} Let $\mX^\eta$ be the interpolation generated by Eq.~\eqref{eq:app-nextgan-particle-update} and compare it with $\bar{\mX}^\eta$ one step at a time. Directly expanding the update and applying the spatial and source-measure Lipschitz bounds gives
\[
\begin{aligned}
\|\mX^{(k+1)}-\bar{\mX}^{(k+1)}\|_{2,N}
&\leq(1+L\eta)\|\mX^{(k)}-\bar{\mX}^{(k)}\|_{2,N}\\
&\quad+\eta\left(\frac1N\sum_{i=1}^N
\left\|\widetilde{\vv}_k(\xx_i^{(k)})-\vv_{\widehat P_k,\widehat Q_M}^{\mu}(\xx_i^{(k)})\right\|^2\right)^{1/2}.
\end{aligned}
\]
Discrete Gronwall therefore gives
\begin{equation}
\sup_{t\in[0,T]}\|\mX^\eta(t)-\bar{\mX}^\eta(t)\|_{2,N}
\leq e^{LT}\eta\sum_{k=0}^{K_T-1}
\left(\frac1N\sum_{i=1}^N
\left\|\widetilde{\vv}_k(\xx_i^{(k)})-\vv_{\widehat P_k,\widehat Q_M}^{\mu}(\xx_i^{(k)})\right\|^2\right)^{1/2}.
\label{eq:app-nextgan-residual-sum}
\end{equation}
Combining Eqs.~\eqref{eq:app-nextgan-stability}--\eqref{eq:app-nextgan-residual-sum} with the triangle inequality proves Eq.~\eqref{eq:app-nextgan-convergence-bound}. Finally, empirical measures formed from i.i.d. samples converge almost surely in $W_2$ under finite second moments, which gives Eq.~\eqref{eq:app-nextgan-convergence-limit}.
\end{proof}

\subsection{Interpretation for neural NextGAN}
\thmref{thm:nextgan-convergence} applies to comparison-space particles driven by a NextGAN velocity oracle. A neural generator introduces an additional map-realization error. If its generated distribution $P_k^{\mathrm{gen}}$ satisfies
\begin{equation}
W_2\!\left(P_{k+1}^{\mathrm{gen}},
\left(\operatorname{Id}+\eta\widetilde{\vv}_k\right)_\#P_k^{\mathrm{gen}}\right)
\leq\gamma_k,
\label{eq:app-generator-realization-error}
\end{equation}
then the same discrete Gronwall argument adds a term bounded by $C_T\sup_k\gamma_k/\eta$ (or by the corresponding weighted sum when the errors are nonuniform) to Eq.~\eqref{eq:app-nextgan-convergence-bound}. Establishing a bound on $\gamma_k$ requires a separate approximation theorem for the generator parameterization and its optimization procedure. Thus the particle result supplies a principled convergence target and an error decomposition for NextGAN, while full neural-training convergence remains outside the present theorem.

\section{Limitations}
\label{app:limitations}

\paragraph{Representation and theoretical scope.}
The distribution-matching characterization in \propref{prop:population-field} concerns the chosen feature spaces and requires a sufficiently rich potential class. Matching these feature distributions does not by itself establish equality of the underlying image or video distributions when the representers discard information. The feature-space ablations in \appref{app:ablations} also show that representer choice affects generation quality. Moreover, the population characterization and particle convergence result do not establish convergence of finite neural critics and generators under alternating optimization; controlling the generator realization error remains an open requirement.

\paragraph{Dependence on pretraining and reference data.}
Our large-scale image and video experiments fine-tune pretrained generators, and the video experiments use LoRA. They therefore do not establish that \NextGAN can train comparably sized generators from scratch. The video reference samples are generated offline by Wan2.1-14B, as described in \appref{app:implementation}. Although training requires no online teacher scores or trajectories, this setup still relies on a pretrained model to construct the reference distribution. Its effectiveness with directly collected real-video training data remains to be evaluated.

\paragraph{Video scaling and evaluation.}
The video results do not show uniform improvements across model sizes and sampling budgets. In \tabref{tab:vbench}, the 14B model remains below the strongest compared baseline in total VBench score at each tested budget. Further controlled experiments are needed to assess the consistency of gains across training runs at this scale.

\end{document}